\documentclass[11pt]{article}
\usepackage{acl}
\usepackage{times}
\usepackage{latexsym}

\usepackage[T1]{fontenc}
\usepackage[utf8]{inputenc}
\usepackage{booktabs}
\usepackage{microtype}
\usepackage{inconsolata}
\usepackage{graphicx}
\usepackage{multirow}
\usepackage{amssymb} 
\usepackage{amsmath}
\usepackage{subcaption}
\usepackage{xspace}
\usepackage{tabularx}
\usepackage{tcolorbox}
\usepackage{subcaption}

\makeatletter
\DeclareRobustCommand\eg{\textit{e.g.,}\@\xspace}

\makeatother
\title{Advancing DialNav through Automatic Embodied Dialog Augmentation}

\author{
\textbf{Leekyeung Han}$^{1}$ \quad
\textbf{Sangwon Jung}$^{2}$ \quad
\textbf{Hyunji Min}$^{1}$ \\
\textbf{Jinseong Jeong}$^{1}$ \quad
\textbf{Minyoung Kim}$^{1}$ \quad
\textbf{Paul Hongsuck Seo}$^{1}$ \\
$^{1}$Korea University \quad
$^{2}$Trillion Labs \\
\texttt{\{happilee12, daream2, dw9030, omniverse186, phseo\}@korea.ac.kr} \\
\texttt{sangwon.jung@trillionlabs.co}
}

\begin{document}
\maketitle
\begin{abstract}
For embodied agents capable of physical interaction, the capability to create and understand dialog is crucial to ensure both safety and effectiveness. 
While DialNav~\cite{han2025dialnav} provides a framework for holistic evaluation of the dialog--execution loop in photorealistic indoor navigation, its performance remains limited by a critical scarcity of training data (2K episodes).
To address this, we propose an automatic generation pipeline, and construct the \textbf{RAINbow} dataset, a large-scale training dataset with 238K episodes for DialNav.
Our pipeline converts existing VLN datasets into multi-turn dialog and creates cost-efficient and high-quality dataset.
Then, we introduce two additional complementary advances to unlock the data's full potential: (1) Dual-Strategy Training, a navigation training scheme to align the navigation training with the dynamic dialog-navigation loop, and (2) a localization model that leverages VLN knowledge. 
By combining these complementary solutions, our model substantially outperforms the baseline in success rate on both \textbf{Val Seen} (58.24, \textbf{+89\%}) and \textbf{Val Unseen} (29.05, \textbf{+100\%}) splits, establishing a new state of the art. 
Our code and dataset are
available at: \href{https://happilee12.github.io/RAINbow}{https://happilee12.github.io/RAINbow}
\end{abstract}

\begin{figure*}[t]
\centering
\includegraphics[width=1\textwidth]{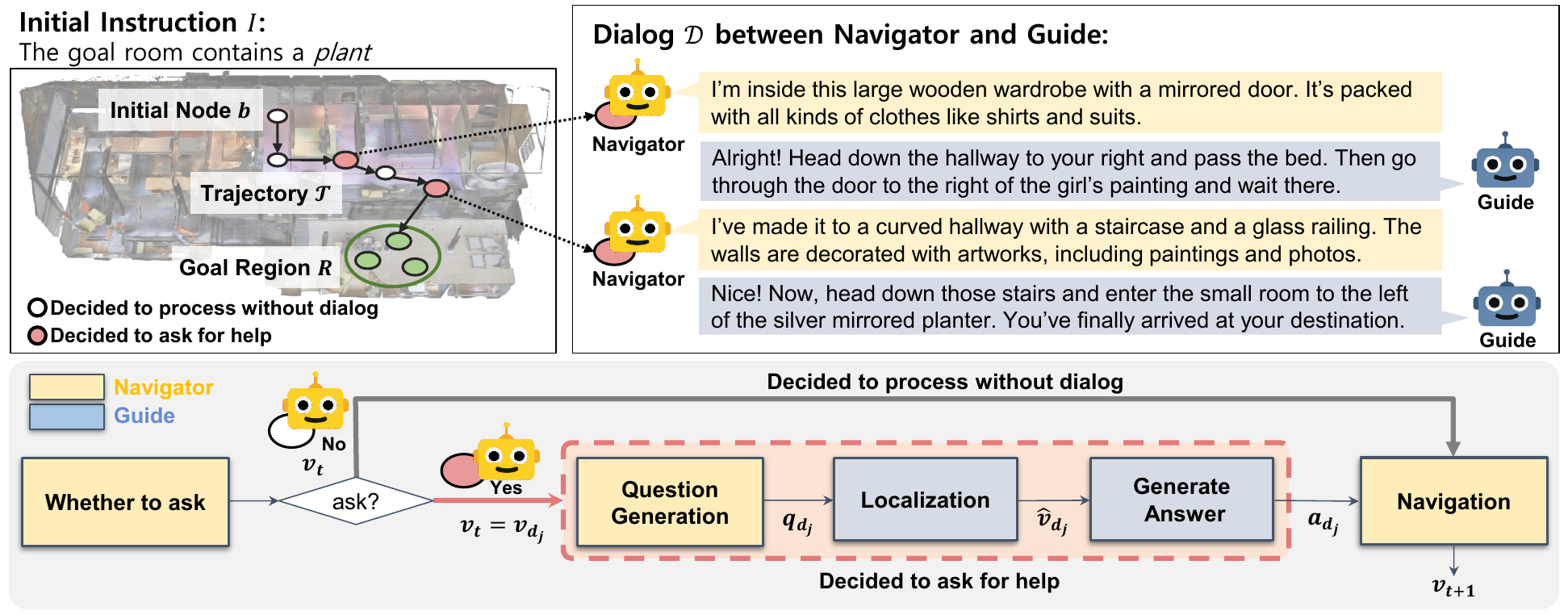}
\caption{\textbf{Overview of the DialNav task}. 
\textbf{Top:} The Navigator starts at an initial node $b$ and navigates to reach the goal region $R$. Since the initial instruction is underspecified, the Navigator engages in multi-turn dialog with a remote Guide to acquire additional guidance along the navigation.  
\textbf{Bottom:} At each step, the Navigator follows a modular decision process: it either proceeds autonomously (\emph{Navigation}) or requests help (\emph{Question Generation}). When a question is asked, the remote Guide localizes the Navigator and provides an answer describing the next path to the goal. This forms an alternating loop of dialog and action that continues until the goal region is reached.  
}
\vspace{-1em}
\label{fig:main}
\end{figure*}

\section{Introduction}

Embodied agents must operate with high reliability, since misinterpreting instructions or executing incorrect actions can cause physical harm. Enabling dialog improves both safety and task effectiveness: by asking questions and refining their understanding before acting, agents can resolve ambiguity and adapt to dynamic situations.
Learning dialog-enabled agents, however, remains highly challenging. Collecting datasets is costly, as it requires two people to coordinate in real time while grounding their conversations in the given task context.
Furthermore, even with such data, training remains difficult because if either the action trajectory or the dialog deviates from the collected annotations, the supervision becomes invalid.

We study these challenges in \textbf{DialNav}~\citep{han2025dialnav}, a cooperative dialog-based vision-and-language navigation (VLN)~\citep{anderson2018vision} task. 
VLN is an embodied navigation problem in which an agent follows a fixed natural language instruction given at the start of an episode to reach a goal. 
DialNav extends this into a dialog-based setup, where navigation unfolds through an interactive dialog exchange.

In DialNav, two agents, a \textit{Navigator} and a remote \textit{Guide}, collaborate through dialog to accomplish the navigation task in a photo-realistic indoor environment (Matterport3D~\citep{chang2017matterport3d}).
The Navigator starts with only an ambiguous and underspecified instruction (\eg ``the target room has a plant''), which is insufficient to reach the goal; success therefore hinges on the following dialog.
To this end, the Navigator proactively asks the Guide for additional guidance during navigation whenever uncertainty arises.
Due to the remote Guide setup in DialNav, the Guide first has to \textit{localize} the Navigator based on this question, which makes the question critical.
The Guide then answers with context-specific instructions that help subsequent navigation decisions. 
To train such agents, \citet{han2025dialnav} first collected the RAIN dataset, a human-human interaction  dataset for DialNav. 
Then, the Navigator and Guide are modeled by separated neural modules specialized for their constituent capabilities (\eg question generation, answer generation, and navigation).
These modules are trained separately on RAIN and later integrated for \textit{holistic} navigation-dialog loop of interactive navigation.

While ~\citet{han2025dialnav} introduced the first dataset and baseline models for DialNav, their performance remains limited. 
A key reason is the severe scarcity of training data: RAIN contains only 2K episodes, which is insufficient to support the high complexity of dialog-based navigation.
Yet collecting human annotated DialNav dataset is exceptionally expensive (\eg 7,500 USD for 2K episodes), as it requires two people to coordinate in real time while grounding their conversations.
Although dataset augmentation techniques to mitigate the cost of  human annotation have been explored in VLN~\citep{fried2018speaker, fan2024navigation, wang2025navrag}, they are restricted to single-turn settings and are not directly applicable to DialNav.

To address this data scarcity challenge, we propose an automatic DialNav episode generation pipeline. Our pipeline leverages existing fine-grained VLN datasets by concatenating their paths, generating scene captions to act as questions, and using an LLM to reformat these into natural dialog. 
Through this process, we yield the \textbf{RAINbow} dataset, which is more than two orders of magnitude larger than the original RAIN.
Despite this substantial data scaling, our preliminary experiments reveal that data augmentation alone is not effective when paired with the existing training framework in \citet{han2025dialnav}. 
We identify that the training scheme from \citet{han2025dialnav}
fails to effectively leverage the augmented data. 
Therefore, we adopt Dual-Strategy Training, a training scheme better aligned with DialNav task that fully exploits our scaled multi-turn dialog-based episodes. 
We additionally improve the localization, an essential subtask to DialNav which remains underexplored.

Together, these advances improve performance of the holistic navigation-dialog loop by more than two-fold: success rate increases from 30.77 to 58.24 (+89\%) on \textit{Val Seen} and from 14.52 to 29.05 (+100\%) on \textit{Val Unseen}. We will publicly release our dataset, code, and models.

In summary, our contributions are threefold:
\begin{itemize}
    \vspace{-0.7em}

    \item We construct RAINbow, a large-scale dataset, expanding the training data by more than two orders of magnitude beyond RAIN.
    \vspace{-0.7em}
    \item We adopt Dual-Strategy Training to better exploit large-scale dialog data and improve localization, an underexplored but critical subtask in DialNav.
    \vspace{-0.7em}
    \item We deliver a new state of the art performance on DialNav, doubling success rate over the previous baseline.
\end{itemize}

\section{Related Work}
\noindent \textbf{Vision and Language Navigation.} \ \ 
Vision-and-Language Navigation (VLN) is an embodied AI task in which an agent navigates through visual environments by following natural language instructions.
Instructions are either coarse-grained~\citep{zhu2021soon,qi2020reverie}, often too ambiguous for reliable navigation, or fine-grained~\citep{anderson2018vision,ku2020room,chen2019touchdown}, often overly detailed and unnatural for human interaction.
Some prior works have explored incorporating dialog or interaction in navigation~\citep{thomason2020vision, roman2020rmm, devries2018talkwalknavigatingnew,banerjee2021robotslang,nguyen2019help,zhu2021self}.
The recently proposed DialNav task~\citep{han2025dialnav} places particular emphasis on the dialog through its remote guide setup and dialog-navigation loop evaluation.
Although DialNav provides a valuable benchmark, its performance is limited primarily due to its small dataset size. 
In this work, we improve DialNav performance by large-scale data augmentation, complemented by enhancements in training.

\noindent \textbf{VLN dataset augmentation.} \
Data scarcity is a well-known challenge in VLN. To address this, various data augmentation approaches have been explored. Some works focused on training a text generation model to generate new instructions for unannotated paths~\citep{fried2018speaker, zhang2023, wang2024}. Other works scaled environments, either by editing existing scenes~\citep{tan2019, liu2021, li2022} or leveraging large sets of additional environments ~\citep{wang2023scaling,Chen_2022_HM3D_AutoVLN,koh2022,guhur2021airbert,lin2023learning}. Recently, more advanced methods have emerged, adopting LLM generators or applying additional filtering or refinement~\citep{zeng2023, fan2024navigation, kong2024, wang2025navrag, wang2024bootstrapping}. 
However, these methods are exclusively focused on generating \textit{single-turn instructions}.
Our work proposes a simple yet powerful pipeline that converts existing single-turn VLN datasets into multi-turn DialNav episodes.
Through this pipeline, we generate RAINbow dataset, which expands the original RAIN dataset~\cite{han2025dialnav} by more than two orders of magnitude.

\noindent \textbf{Embodied Dialog.} \ \ 
Embodied dialog presents challenges in both dataset and training.
To mitigate the challenge of data scarcity, some works explore data augmentation leveraging large language models~\citep{padmakumar-etal-2023-multimodal} or template-based procedures~\citep{gao2022dialfred}. 
For training, many of existing works primarily train and evaluate models on static dialog histories~\citep{thomason2020vision,padmakumar2022teach,hahn2020you}. 
Some works propose training scheme for holistic execution-dialog loop setups, using human correction~\citep{devries2018talkwalknavigatingnew}, reinforcement learning~\citep{roman2020rmm}, or training with all possible dialog cases~\citep{gao2022dialfred}. 
The prior work on DialNav~\citep{han2025dialnav} trains agents in a static setup, which creates a train-test mismatch.
In this work, motivated by recent findings that large-scale synthetic training data can substantially improve model capabilities~\cite{wang2024bootstrapping,min2025goat}, we generate a high-quality, well-grounded dataset at low cost, and adopt a training scheme to fully utilize this large-scale data.

\vspace{-0.1cm}
\section{Preliminaries}
\vspace{-0.2cm}
\noindent \textbf{Notation.} \ \   
In DialNav~\cite{han2025dialnav}, the \textit{Navigator} agent traverses toward a goal with the assistance of a remote \textit{Guide}, as illustrated in Figure \ref{fig:main} (top).
Formally, the environment is represented as a connectivity graph $G=(V,E)$, where $v\in V$ is a navigable node, and $E \subseteq V \times V$ is the set of navigable edges between nodes.
A DialNav episode $\mathcal{E}$ is defined as $\mathcal{E} = (G, b, R, I, T_J, D_J)$, where $b\in V$ is the initial node, $R\subseteq V$ is the goal region spanning multiple adjacent nodes, and $I$ is an initial instruction. 
The Navigator's complete trajectory $\mathcal{T}$ is a sequence of traversed nodes over $K$ steps $\mathcal{T} = (v_0, v_1, \ldots, v_K)$, where $v_0=b$ and $v_K \in R$. 
During navigation, $J$-turn dialog occurs, denoted as $D_J = ((q_1, a_1), \ldots, (q_{J}, a_{J}))$, with each $(q_j, a_j)$ exchanged at a dialog point $v_{d_j} \in \mathcal{T}$.
These $J$ dialog points segment the full trajectory $\mathcal{T}$ into $J+1$ sub-trajectories: $(T^{(0)}, T^{(1)}, \ldots, T^{(J)})$, where $T^{(j)}=(v_{d_j}, \ldots, v_{d_{j+1}-1})$.

\begin{figure*}[t]
\centering
\includegraphics[width=1\textwidth]{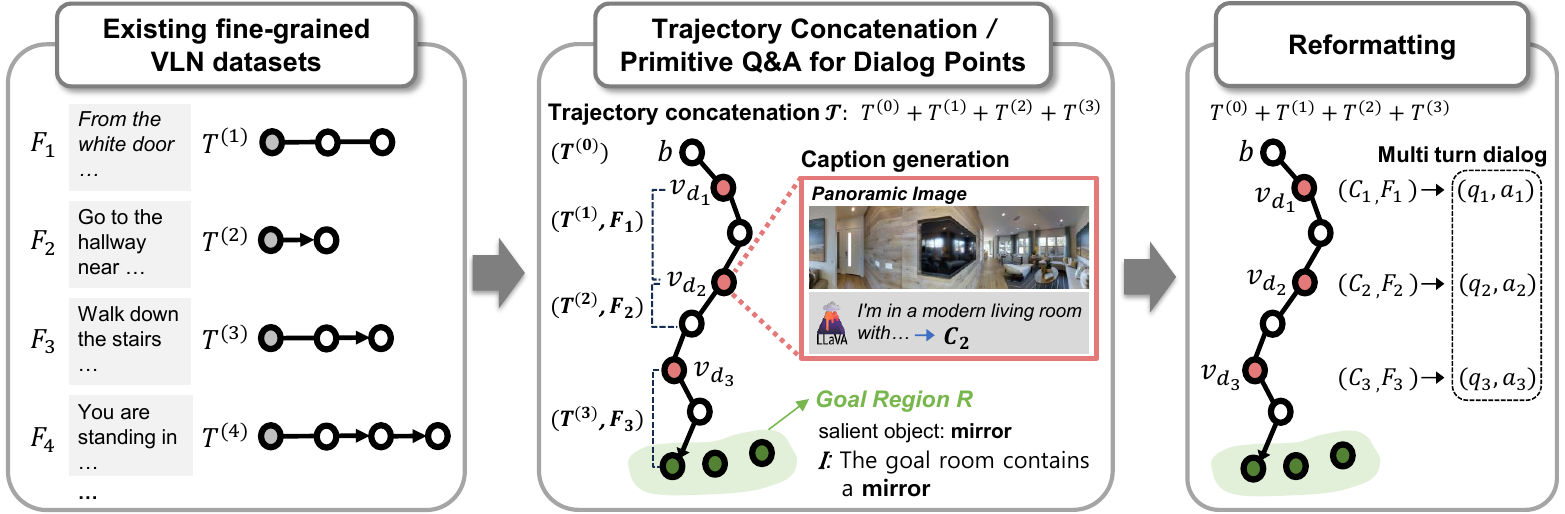}
\caption{
\textbf{Overview of the dataset generation pipeline.} 
\textbf{(Left)} We start from existing single-turn fine-grained VLN datasets, where each path is paired with its instruction $F_j$. 
\textbf{(Middle)} Multiple sub-trajectories are concatenated into an extended trajectory. The starting node of each sub-trajectory becomes a dialog point $v_{d_j}$, and at each dialog point, a panoramic caption $C_j$ is generated using a vision–language model. The original fine-grained instructions $F_j$ are repurposed as dialog answers, while the final node defines the goal region $R$. 
\textbf{(Right)} Caption–instruction pairs $(C_j, F_j)$ are then reformatted into natural multi-turn dialogs using a large language model, producing large-scale dialog-style data for DialNav training.
}
\label{fig:data_construction}
\vspace{-1.5em}

\end{figure*}
        
\noindent \textbf{Navigator and Guide Workflow.} \ \ 
The DialNav task unfolds as a holistic workflow integrating all subtasks (Figure \ref{fig:main} bottom).
The episode begins with the Navigator at $b$ with the initial instruction $I$, and the remote Guide with knowledge of the full environment $G$, the goal $R$, and $I$.
The process then unfolds in rounds.
At each step $t$, the Navigator first decides whether to ask for help.
If the Navigator chooses to \textit{proceed without help}, it autonomously selects its next action and advances to node $v_{t+1}$.
If the Navigator chooses to \textit{ask for help}, it generates a natural language question $q_j$, and its current node is denoted as a dialog point $v_{d_j}$.
Upon receiving the question, the remote Guide first localizes the Navigator by estimating its dialog point as $\hat{v}_{d_j}$. 
Then, it generates a natural language answer $a_j$, typically describing the path from the inferred position $\hat{v}_{d_j}$ to $R$.
The Navigator continues navigation using this exchange $(q_j, a_j)$ as additional guidance. 
This process repeats until the Navigator determines it has reached the goal $R$.

\section{Method}
\vspace{-0.2cm}

In this section, we first present our automatic episode generation pipeline (\S\ref{sec:dataset}). 
This pipeline leverages existing VLN datasets to synthesize a large-scale dataset, \textbf{RAINbow}, expanding the available training data by over two orders of magnitude. 
We then introduce complementary enhancements to the navigation training (\S\ref{sec:nav-training}) and localization model (\S\ref{sec:loc-model}), enabling the model to better exploit this large-scale augmented dataset.

\begin{figure*}[t]
\centering
\includegraphics[width=1\textwidth]{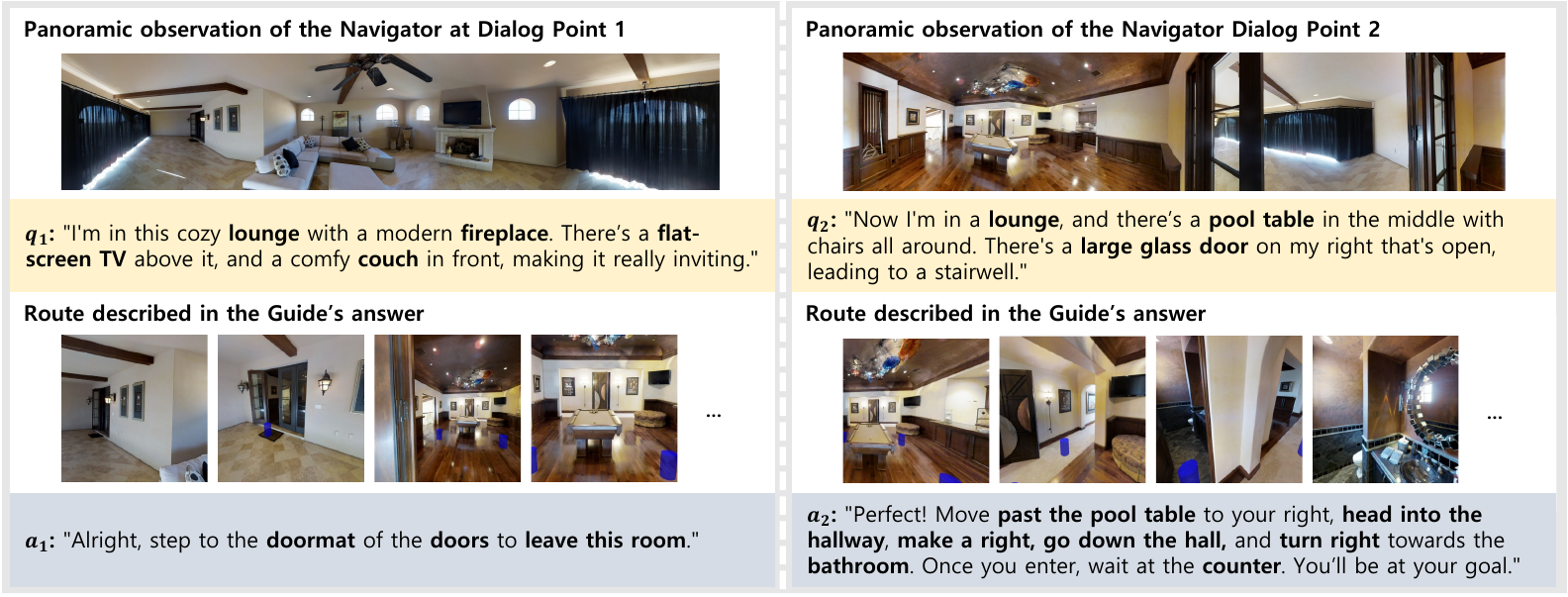}
\caption{\textbf{Qualitative example of RAINbow.} This figure shows a 2-turn dialog episode in RAINbow. The left column shows the first dialog exchange, and the right column shows the second. The generated dialogs exhibit a natural conversational flow across turns and are well-grounded in both the Navigator's visual observations (\eg ``fireplace'', ``pool table'') and the route the Guide describes (\eg ``leave this room'', ``past the pool table'').}
\vspace{-1.5em}
\label{fig:qualitative_data}
\end{figure*}

\vspace{-0.7em}
\subsection{RAINbow Dataset Generation}
\label{sec:dataset}
\vspace{-0.5em}

Collecting dialog-based navigation datasets is prohibitively expensive, as it requires two people to coordinate in real time while grounding their conversations in the given task context.
Consequently, the human-annotated RAIN dataset~\cite{han2025dialnav} contains only 2K episodes, which is insufficient to capture the complexity of dialog-conditioned navigation and thus hinders generalization.
To overcome this limitation, we automatically generate large-scale DialNav episodes by utilizing VLN datasets. 
We observe that, in DialNav~\cite{han2025dialnav}, dialog unfolds in turns along a long trajectory, where each question--answer exchange guides a subsequent portion of the path. To simulate this characteristic, we designe our pipeline in three stages: (1) concatenating paths from existing VLN datasets; (2) generating scene captions for each dialog point (to serve as question content); and (3) rewriting the caption-instruction pairs into natural dialog using an LLM. 
Figure~\ref{fig:data_construction} illustrates the RAINbow data generation pipeline.
We describe each stage in detail below.

\noindent \textbf{Trajectory Concatenation.} \ \
\label{sec:traj_concat}
\label{sec:trajectory_concatenation}
From three widely used VLN datasets, R2R~\citep{anderson2018vision}, RxR~\citep{ku2020room} and CVDN\footnote{Although CVDN is a dialog dataset, its answers describe the 5 next nodes along the shortest path to the goal, which we utilize as fine-grained instructions.}
~\citep{thomason2020vision}, we concatenate 2--4 paths to form a new full trajectory $\mathcal{T}$.
Each path, which becomes a sub-trajectory $T^{(j)}$, is paired with its original fine-grained instruction $F_j$.
We then form the initial exploration path $T^{(0)}$ by prepending up to five navigation steps to the start node of $T^{(1)}$.
To better reflect the noise encountered in real dialog-based navigation, we additionally inject \emph{mislocalization}, \emph{misnavigation}, and \emph{exploration} scenarios into 10\% of the constructed trajectories and their corresponding dialogs.
From the resulting trajectory $\mathcal{T} = (T^{(0)}, \ldots, T^{(J)})$, the initial node of $T^{(0)}$ becomes the episode's start node $b$, and the region containing the final node of $T^{(J)}$ becomes the goal region $R$.
We then generate an initial instruction $I$ that indirectly specifies the goal region $R$ by describing an object located within it.
This region and object information is based on the metadata in~\citet{chang2017matterport3d}.
Further implementation details are provided in Supp. Mat.~\ref{supp:traj_concat}.

\vspace{-0.3em}
\noindent \textbf{Primitive Q\&A for Dialog Points.} 
\label{sec:gen_qa}
Given the concatenated trajectory $T$, the initial node $v_{d_j}$ of each sub-trajectory $T^{(j)}$(for $j \ge 1$) is treated as a \emph{dialog point} where a dialog exchange occurs. 
In this step, we prepare the primitive content for the dialog exchange ($q_j$, $a_j$) at each dialog point $v_{d_j}$.
Leveraging the question style of RAIN~\cite{han2025dialnav}, where questions typically describe the visual scene to aid localization, we first generate the question content, $C_j$, by captioning the observation at the dialog point via a large vision–language model (e.g., LLaVA-1.5-7B~\citep{liu2023llava}).
For the answer content, we repurpose the original fine-grained instruction $F_j$ associated with the sub-trajectory $T^{(j)}$.
The resulting $(C_j, F_j)$ thus form primitive question--answer pairs aligned with the dialog point of the concatenated trajectory. 
As these raw pairs are not yet in natural dialog form and the flow across turns is often awkward, we refine them in the following step. 
Details are provided in the Supp. Mat.~\ref{supp:question_generation}.

\vspace{-0.2em}

\noindent \textbf{Reformatting into Multi-turn Dialog.} \ \ 
\label{sec:reformat}
Finally, we reformat the caption–instruction pairs $(C_j, F_j)$ obtained from the previous steps into coherent multi-turn dialogs. 
While the caption $C_j$ describes the local scene and the instruction $F_j$ gives a specific direction for the next sub-trajectory $T^{(j)}$, these sentences are originally written as independent sentences and therefore do not form a natural conversation.
To bridge this gap, we employ an LLM (\eg GPT-4o-mini~\citep{openai2024gpt4omini}) to rewrite the raw $(C_j, F_j)$ pairs into fluent question–answer exchanges $(q_j, a_j)$.
To reduce hallucination and ensure faithfulness to the intended goal, we first refine each instruction before converting it into dialog, and further validate this process through human evaluation and closed-source VLMs.
Details of the prompting and validation are provided in the Supp. Mat.~\ref{supp:reformat}, ~\ref{supp:dataset_evaluation}, ~\ref{supp:hallucination-analysis}.  

\vspace{-0.3em}
\noindent \textbf{The RAINbow Dataset.} \ \ 
Through this process, we synthesize the \textbf{RAINbow} (\textbf{RAIN} \textbf{b}uilt \textbf{o}n \textbf{w}ide set) dataset.
The resulting RAINbow is a 238K episode dataset that is over two orders of magnitude larger than the RAIN dataset.
Figure~\ref{fig:qualitative_data} shows an example of the final generated dialog data.
The key characteristics in RAINbow remain comparable to the RAIN dataset. In Rainbow, the average numbers of dialog turns, trajectory nodes, question words, and answer words are 2.71, 19.54, 30.43, and 43.17, respectively (compared to 1.88, 26.03, 27.48, and 42.53 in RAIN).
The reformatting step using GPT-4o-mini incurred a total cost of about USD 400, equating to USD 0.0016 per episode.
This makes our generation pipeline about 2,000 times more cost-effective than manual annotation in RAIN (USD 3.75 per episode).
Statistics and more examples of RAINbow are provided in Supp. Mat.~\ref{supp:dataset_stats}.

\vspace{-0.5em}
\subsection{Training the Navigator and Guide}
\vspace{-0.3em}
To train the Navigator and the Guide, we follow the modular design introduced in \citet{han2025dialnav}. 
As described in the workflow in Figure~\ref{fig:main} (bottom), each agent is decomposed into task-specific modules—\textit{whether-to-ask}, \textit{question generation}, and \textit{navigation} for the Navigator, and \textit{localization} and \textit{answer generation} for the Guide.
For the whether-to-ask module, we adopt a confidence thresholding strategy, prompting the agent to initiate a question whenever the model confidence falls below a predefined threshold.
For the question and answer generation modules, we follow the training scheme of \citet{han2025dialnav}.

\vspace{-0.4em}
\subsubsection{Dual-Strategy Training for Navigation}
\vspace{-0.3em}
\label{sec:nav-training}
The segment-based training in prior work~\cite{han2025dialnav} trains the navigator without previous navigation or future dialog updates, not taking the evolving dialog nature of DialNav into account.
While training on the full episode with dialog updates along the path seems to solve this, it presents a fundamental dilemma: the agent must strictly follow the annotated path to receive valid dialog updates $(q_j, a_j)$ at the correct nodes $v_{d_j}$.
However, this strict adherence prevents the agent from learning to recover from its own errors.

\begin{figure}[t]
\centering
\begin{subfigure}[b]{0.7\linewidth}
    \centering
    \includegraphics[width=\linewidth]{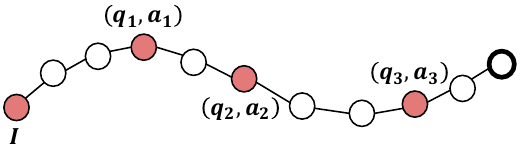}
    \vspace{-0.5cm}
    \caption{DialNav training episode $\mathcal{E}$}
\label{fig:episode_data}
\end{subfigure}
\begin{subfigure}[b]{0.7\linewidth}
    \centering
    \includegraphics[width=\linewidth]{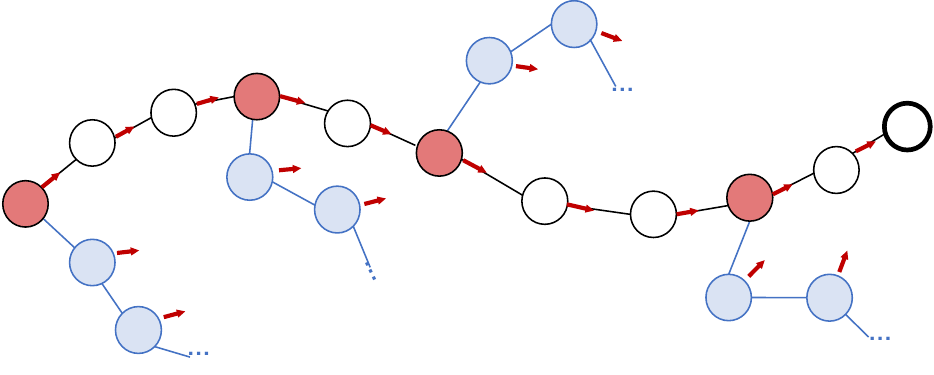}
    \vspace{-0.3cm}
    \caption{Dual-Strategy Training}
\label{fig:episodic_training}
\end{subfigure}
\caption{\textbf{Comparison of navigation training strategies.}
(a) \textbf{DialNav training episode $\mathcal{E}$}: An episode unfolds from an initial node (marked $I$), with dialog dynamically updated at each dialog points (red nodes).
(b) \textbf{Dual-Strategy Training}: The\textit{ data-guided rollout}, follows the annotated path in dataset (white nodes), updating dialog at each dialog point (red nodes). This is supplemented with \textit{on-policy rollouts} (blue nodes) forked at each dialog point.
The model is supervised to predict the next node along the dataset path for data-guided rollouts and the shortest path to the goal for on-policy rollouts, respectively (red arrows).}

\label{fig:episodic-training}
\vspace{-1.9em}
\end{figure}

To this end, we adopt a \textbf{Dual-Strategy Training} for navigation that combines two types of rollouts: data-guided and on-policy (Figure~\ref{fig:episodic_training}).
The \textit{data-guided rollout} (Figure~\ref{fig:episodic_training}, white nodes) constrains the Navigator to follow the dataset trajectory, allowing it to arrive at each dialog node $v_{d_j}$ and receive the ground-truth dialog exchange $(q_j, a_j)$ as a dynamic instruction update.
This is complemented by an \textit{on-policy rollout} (Figure~\ref{fig:episodic_training}, blue nodes), forked at each dialog point $v_{d_j}$ (Figure~\ref{fig:episodic_training}, red nodes), where the agent follows its own policy without receiving further dialog updates, supervised to predict the next node on the shortest path to the goal region $R$.
(Figure~\ref{fig:episodic_training}, red arrows).
By jointly training on both data-guided and on-policy rollouts, the navigation module learns to leverage full episode while remaining robust to its own prediction errors.

The loss for each dialog point, $v_{d_j}$ is defined as followings:
\begin{equation}
\text{
\small  $ {
\begin{aligned}
    \mathcal{L} = & \underbrace{
        \sum_{i=d_j}^{d_{j+1}-1} \log p(v_{i+1}|v_{1:i}, D_j)
        }_{\text{data-guided rollout}} \\
       & +
    \underbrace{
        \sum_{t=d_j}^{K-1} \log p(\hat{v}_{t+1}|v_{1:d_j},\bar{v}_{d_j+1:t},  D_j)
       }_{\text{on-policy rollout}},
\end{aligned}
    } $}
\end{equation}
in which $\hat{v}$ is the next step on the shortest path and  $\bar{v}$ denotes the next step sampled the own policy of the navigation model.



\subsubsection{Transferring VLN Knowledge for Localization}
\label{sec:loc-model}
Localization is a subtask arising from DialNav’s remote Guide setup.
The Guide, who knows the entire map $G$, must predict the dialog point $v_{d_j}$ based on the Navigator's question $q_j$ to provide accurate guidance.
Localization tasks share similarities with VLN in its environment and task, as both operate in the Matterport3D~\cite{chang2017matterport3d} and are text-grounded node selection tasks.
This suggests that localization can benefit from models and pretraining strategies originally developed for VLN~\cite{hahn2022transformer}.
Motivated by this, we adopt the graph-based Transformer architecture of \citet{chen2022think} for localization.
As shown in our experiments, this significantly improves localization performance.
Details of the model are provided in Supp. Mat. \ref{sec:model-detail}.

\setlength{\tabcolsep}{5pt}
\begin{table*}[t]
    \centering

    \scalebox{0.77}{
        \begin{tabular}{r l cccccc cccccc}
        \toprule
        \textbf{\#} & \textbf{Setup} &
        \multicolumn{6}{c}{\textbf{Val Seen}} &
        \multicolumn{6}{c}{\textbf{Val Unseen}} \\
        \cmidrule(lr){3-8} \cmidrule(lr){9-14}
        & & SR$\uparrow$ & OSR$\uparrow$ & SPL$\uparrow$ & NE$\downarrow$ & NSC & DTC
        & SR$\uparrow$ & OSR$\uparrow$ & SPL$\uparrow$ & NE$\downarrow$ & NSC & DTC \\
        \midrule
        (1) & Baseline~\cite{han2025dialnav} & 30.77 & 38.46 & 26.93 & 9.76 & 18.65 & 3.44
        & 14.52 & 19.50 & 11.33 & 16.38 & 15.52 & 5.61 \\
        (2) & +RAINbow & 34.07 & 43.96 & 32.21 & 9.61 & 19.64 & 5.34
        & 18.26 & 21.58 & 13.89 & 16.27 & 23.00 & 13.76 \\
        (3) & +DST & 38.46 & 45.05 & 32.78 & 8.92 & 22.58 & 6.38
        & 15.77 & 24.90 & 11.79 & 16.92 & 21.29 & 10.16 \\
        (4) & +RAINbow+DST & 50.55 & 60.44 & 45.60 & 6.22 & 22.13 & 5.63
        & 25.73 &\textbf{ 32.78} & 19.13 & 14.49 & 23.85 & 13.30 \\
        (5) & \textbf{+RAINbow+DST+GTL (Ours)}
        & \textbf{58.24} & \textbf{68.13} & \textbf{51.65} & \textbf{5.38}& 21.30 & 4.66
        & \textbf{29.05} & 31.54 & \textbf{19.61} & \textbf{13.91} & 21.46 & 11.08 \\
        \bottomrule
        \end{tabular}
    }
    \vspace{-0.5em}
    \caption{\textbf{Performance gains from progressively adding components.} \textbf{+RAINbow}: training with RAINbow dataset (\S\ref{sec:dataset}). \textbf{+DST}: training navigation module with Dual-Strategy Training (\S\ref{sec:nav-training}). \textbf{+GTL}: adopting Graph-based Transformer Localization model (\S\ref{sec:loc-model}). RAINbow alone (Row 2) offers performance gains, but its potential is unlocked when combined with our proposed DST (Row 4), which significantly amplifies its effect. GTL (Row 5) provides further improvement, leading to the best overall performance.
    }
    \vspace{-0.2cm}
    \label{tab:holistic-results}
\end{table*}
\setlength{\tabcolsep}{6pt}


\setlength{\tabcolsep}{5pt}
\begin{table*}[t]
    \centering

    \scalebox{0.8}{
        \begin{tabular}{r l cccccc cccccc}
        \toprule
        \textbf{\#} & \textbf{Dataset} & 
        \multicolumn{6}{c}{\textbf{Val Seen}} &
        \multicolumn{6}{c}{\textbf{Val Unseen}} \\
        \cmidrule(lr){3-8} \cmidrule(lr){9-14}
        & & SR$\uparrow$ & OSR$\uparrow$ & SPL$\uparrow$ & NE$\downarrow$ & NSC & DTC
          & SR$\uparrow$ & OSR$\uparrow$ & SPL$\uparrow$ & NE$\downarrow$ & NSC & DTC \\
        \midrule
        (1) & RAIN   & 37.36 & 47.25 & 33.74 & 8.54 & 22.16 & 6.01 
              & 23.24 & 29.46 & 16.49 & 14.68 & 20.80 & 8.85 \\
        (2) & + VLN instructions & 40.66 & 51.65 & 32.73 & 8.94 & 22.93 & 9.56 
              & 25.73 & \textbf{34.44} & 17.92 & 14.37 & 24.68 & 12.67 \\
        (3) & + Traj. Concat  & 39.56 & 48.35 & 35.36 & 7.46 & 20.54 & 4.00
              & 19.09 & 27.80 & 13.79 & 17.04 & 25.80 & 13.06 \\
        (4) & + Primitive Q\&A  & 56.04 & 65.93 & 51.41 & 5.51 & 20.22 & 5.80 
              & 22.82 & \textbf{34.44} & 15.47 & 14.32 & 23.15 & 12.19 \\
        (5) & \textbf{+ Reformat (RAINbow)} & \textbf{58.24} & \textbf{68.13} & \textbf{51.65} & \textbf{5.38} & 21.30 & 4.66 
              & \textbf{29.05} & 31.54 & \textbf{19.61} & \textbf{13.91} & 21.46 & 11.08 \\
        \bottomrule
        \end{tabular}
    }
    \vspace{-0.5em}
    \caption{\textbf{Dataset augmentation ablations.} 
        RAIN: training only with the RAIN dataset.
        \textbf{+ VLN instructions}: adding VLN instructions. 
        \textbf{+ Traj. Concat}: concatenating trajectories from VLN instructions. 
        \textbf{+ Primitive Q\&A}: using primitive QA content. 
        \textbf{+ Reformat}: reformatting into natural multi-turn QA dialogs. 
    }
\vspace{-1.5em}
\label{tab:dataset_ablation}
\end{table*}
\setlength{\tabcolsep}{6pt}

\vspace{-0.5em}
\section{Experiments}
\vspace{-0.5em}
\subsection{Experimental Settings}
\noindent \textbf{Implementation of Navigator and Guide.} \ \
Following prior work~\cite{han2025dialnav}, we use DUET~\citep{chen2022think} for \textit{navigation} and LANA~\citep{wang2023lana} for the \textit{question/answer generation} modules.
For the \textit{localization} module, we compare the GCN from~\citet{han2025dialnav} with Graph-based Transformer Localization (\S \ref{sec:loc-model}).
All models are initialized from publicly available pre-trained weights.
We then train the modules using a 1:9 mixture of RAIN and RAINbow data.
For the \textit{whether-to-ask} module, we fix the confidence threshold of $0.9$.
Details of each model architecture are provided in Supp. Mat.~\ref{sec:model-detail}.

\noindent \textbf{Evaluation.} \ \  
We evaluate our method on the RAIN validation dataset, which is split into \textit{Val Seen} and \textit{Val Unseen}~\citep{anderson2018vision}. 
The \textit{Val Seen} split consists of environments seen during training, albeit with novel paths, whereas \textit{Val Unseen} contains entirely novel environments.
We report performance under the holistic dialog–navigation loop setting, where the Navigator starts with an initial instruction and must proactively ask questions to receive guidance and reach the goal.
We measure performance using standard VLN metrics: Success Rate (SR), Oracle Success Rate (OSR), Success weighted by Path Length (SPL), and Navigation Error (NE, in meters), which capture navigation quality and efficiency. We also report Navigation Step Count (NSC) and Dialog Turn Count (DTC) to measure exploration behavior and dialog usage.

\vspace{-0.5em}
\subsection{Results}
\vspace{-0.3em}
\noindent \textbf{Impact of the proposed components.} \ \
Table~\ref{tab:holistic-results} presents the impact of our proposed components: adding the RAINbow dataset, Dual-Strategy Training (DST), and Graph-based Transformer Localization (GTL).
RAINbow alone (Row~2) provides a modest improvement over the baseline (Row~1) in SR under both setups (Val Seen +3.30, Val Unseen +3.74). 
However, its true potential is realized when paired with DST (Row~4),  which yields substantially larger gains (Val Seen +12.09, Val Unseen +9.96, compared to Row~3). 
This result confirms our hypothesis: while the baseline segment-based training~\cite{han2025dialnav} cannot fully leverage RAINbow, DST effectively utilizes the large-scale, multi-turn episodes and unlocks their full potential.
Finally, adding GTL (Row~5) produces the best performance: SR reaches 58.24 (+89\%) on Val Seen and 29.05 (+100\%) on Val Unseen, roughly doubling the baseline success rate on both splits.

\noindent \textbf{Ablations of the Data Augmentation Pipeline.} 
Table~\ref{tab:dataset_ablation} analyzes the effect of each stage in our augmentation pipeline.
For fair comparison, we applied Dual-Strategy Training and Graph-based Transformer Localization across all setups.
Naively adding publicly available fine-grained VLN instructions~\cite{anderson2018vision, ku2020room} (Row~2), providing a baseline for data augmentation, improved navigation SR +3.30~(+9\%) in Val Seen and +2.69~(+11\%) in Val Unseen. 
This gain from simply exposing the Navigator to more environments highlights the importance of data diversity for improving performance.
With trajectory concatenation (Row~3), the performance drops compared to Row~2.
We attribute this to data reduction; the constraints for connecting paths (\S\ref{sec:traj_concat}) necessarily discard some source trajectories, providing the navigator with less environmental coverage than the full VLN instruction set.
For Rows~2 and 3, only the navigation module can use the augmented data, because these settings lack the dialog content required to train the localization, question generation, and answer generation modules.
Introducing Primitive Q\&A (Row~4) addresses this issue by providing supervision across all modules, resulting in a substantial improvement in SR on Val Seen: +18.68 (+41.66\%) compared to Row~1.
However, this setting still struggles in  Val Unseen. 
We conjecture this is due to a style mismatch, as modules are trained on a mixture of human-annotated RAIN dialogs and the unformatted, raw content from the pipeline, which hinders generalization.
Finally, Reformat (Row~5) converts this raw content into natural, contextually coherent dialogs, yielding the best results (Seen SR: 58.24, Unseen SR: 29.05).
Crucially, reformatting significantly boosted SR in Val Unseen (compared to Row~4, +6.23). 
Overall, these results demonstrate that while raw VLN fine-grained data or naively concatenated trajectories provide limited benefits, our pipeline's full process of structuring and naturalizing content into consistent dialog is effective for DialNav.



\setlength{\tabcolsep}{5pt}
\begin{table}[t]
    \centering
    \scalebox{0.8}{
    \begin{tabular}{r l cccc}
    \toprule
    \textbf{\#} & \textbf{Training Method} & SR$\uparrow$ & SPL$\uparrow$ & NSC & DTC \\
    \midrule
    \multicolumn{6}{l}{\textit{Val Seen}} \\
    (1) & \ \ \citet{han2025dialnav} & 32.97 & 30.76 & 19.85 & 5.76 \\
    (2) & \ \ \textbf{DST} & \textbf{58.24} & \textbf{51.65} & 21.30 & 4.66 \\
    (3) & \ \ \ \ -- On-policy Rollout & \textbf{58.24} & 50.05 & 20.80 & 6.21 \\
    \midrule
    \multicolumn{6}{l}{\textit{Val Unseen}} \\
    (4) & \ \ \citet{han2025dialnav} & 22.82 & 17.00 & 20.89 & 10.57 \\
    (5) & \ \ \textbf{DST} & \textbf{29.05} & \textbf{19.61} & 21.46 & 11.08 \\
    (6) & \ \ \ \ -- On-policy Rollout & 21.16 & 15.38 & 21.41 & 8.34 \\
    \bottomrule
    \end{tabular}
    }
    \caption{\textbf{Navigation training methods ablation.} Comparison of baseline and Dual-Strategy Training (DST) with and without On-policy Rollout. }
    \label{tab:training-ablation-full} 
    \vspace{-0.3cm}
\end{table}
\setlength{\tabcolsep}{6pt}

\begin{table}[t]
\centering
\scalebox{0.8}{
\begin{tabular}{lccc}
\toprule
\textbf{Detour ratio} & $1.0\text{--}1.3$ & $\geq 1.3$ & Total \\
\midrule
\textbf{DST} & 11.20 & \textbf{17.85} & \textbf{29.05} \\
-- On-policy Rollout & 10.79 & 10.37 & 21.16 \\
\bottomrule
\end{tabular}
}
\caption{\\textbf{Detour analysis on Val Unseen.} Comparison of Success Rate between DST and the variant without on-policy rollout across different detour ratio. }

    \label{tab:detour_sr}
    \vspace{-0.3cm}
\end{table}

\noindent \textbf{Ablations of Navigation Training.} \
Table~\ref{tab:training-ablation-full} ablates Dual-Strategy Training (\S\ref{sec:nav-training}) on both the Val Seen and Val Unseen splits.
For fair comparison, we applied the RAINbow data and Graph-based Transformer Localization across all setups.
Comparing Dual-Strategy Training (DST) (Row 2 and 5) to the Baseline (Row 1 and 4), we see a dramatic improvement in SR across both splits (Val Seen +76.6\%, Val Unseen +27.3\%). 
This confirms that Dual-Strategy Training (DST) that leverages the full, dynamic dialog loop is critical. 
Row 3 and 6 ablates our training by removing the on-policy rollout. 
While this maintains high SR in Val Seen, there is a severe drop in Val Unseen (-7.89).
Table~\ref{tab:detour_sr} further analyzes SR by detour ratio (path length / shortest path) on Val Unseen. 
When the detour is small (1.0--1.3), DST and the variant without on-policy rollout achieve comparable SR (11.20 vs.\ 10.79). 
However, when the detour is larger ($\geq$ 1.3), DST yields substantially higher SR (17.84) than the variant without on-policy rollout (10.37). 
By explicitly incorporating exploration through on-policy rollout during training, our model learns to recover from trajectory errors and make more robust predictions.



\setlength{\tabcolsep}{5pt}
\begin{table}[t]
    \centering
    \scalebox{0.8}{
        \begin{tabular}{r c cc cc cc}
        \toprule
        \multirow{2}{*}{\textbf{\#}} & \multirow{2}{*}{\textbf{AC}} & \multirow{2}{*}{\textbf{VT}}   & \multicolumn{2}{c}{\textbf{Val Seen}} & \multicolumn{2}{c}{\textbf{Val Unseen}} \\
        \cmidrule(lr){4-5} \cmidrule(lr){6-7}
        & & & LE$\downarrow$ & A@3m$\uparrow$ & LE$\downarrow$ & A@3m$\uparrow$ \\
        \midrule
        (1) & & & 10.72 & 40.70 & 14.47 & 21.74 \\
        (2) & \checkmark & & 17.42 & 16.37 & 16.25 & 15.31 \\
        (3) & \checkmark & \checkmark & \textbf{8.16} & \textbf{49.12} &\textbf{ 11.45} & \textbf{34.97} \\
        \bottomrule
        \end{tabular}
    }
    \caption{\textbf{Localization model ablation.} \textbf{AC}: Architecture change to Graph-based Transformer Localization. \textbf{VT}: VLN pretrained knowledge transfer. All together, the new localization model reduces localization error (LE) by 2.56m in Val Seen and 3.02m in Val Unseen compared to the baseline.}
    \vspace{-0.5cm}
    \label{tab:localization-ablation}
\end{table}
\setlength{\tabcolsep}{6pt}


\setlength{\tabcolsep}{5pt}
\begin{table}[t]
    \centering
    \scalebox{0.8}{
        \begin{tabular}{l c cc cc}
        \toprule
         & & \multicolumn{2}{c}{\textbf{Val Seen}} & \multicolumn{2}{c}{\textbf{Val Unseen}} \\
        \cmidrule(lr){3-4} \cmidrule(lr){5-6}
         \textbf{Agent Role} & \textbf{Method} & SR $\uparrow$ & HS $\uparrow$ & SR $\uparrow$ & HS $\uparrow$ \\
        \midrule
        \multirow{2}{*}{\textit{Navigator}} & Baseline &39.1  & 2.89 & 13.6 & 2.16 \\
        & \textbf{Ours} & \textbf{43.5} & \textbf{3.18} & \textbf{20.0} & \textbf{2.24} \\
        \midrule
        \multirow{2}{*}{\textit{Guide}} & Baseline & 47.8 & 3.18 & 14.3 & 2.10 \\
        & \textbf{Ours} & \textbf{73.1} & \textbf{3.75} & \textbf{44.0} & \textbf{2.52} \\
        \bottomrule
        \end{tabular}
    }
    \caption{\textbf{Human-agent cooperation results on DialNav.} \textbf{SR}: Navigation Success Rate (\%). \textbf{HS}: Human Score, representing the perceived helpfulness of the agent counterpart, rated by human participants on a scale from 1 to 5. For each setting, either the human or the agent takes the Guide role, while the other takes the Navigator role.}
    \vspace{-0.5cm}
    \label{tab:human-eval-final}
\end{table}
\setlength{\tabcolsep}{6pt}
\begin{figure*}[t]
    \centering 

    \begin{subfigure}[b]{0.49\linewidth}
        \includegraphics[width=\linewidth]{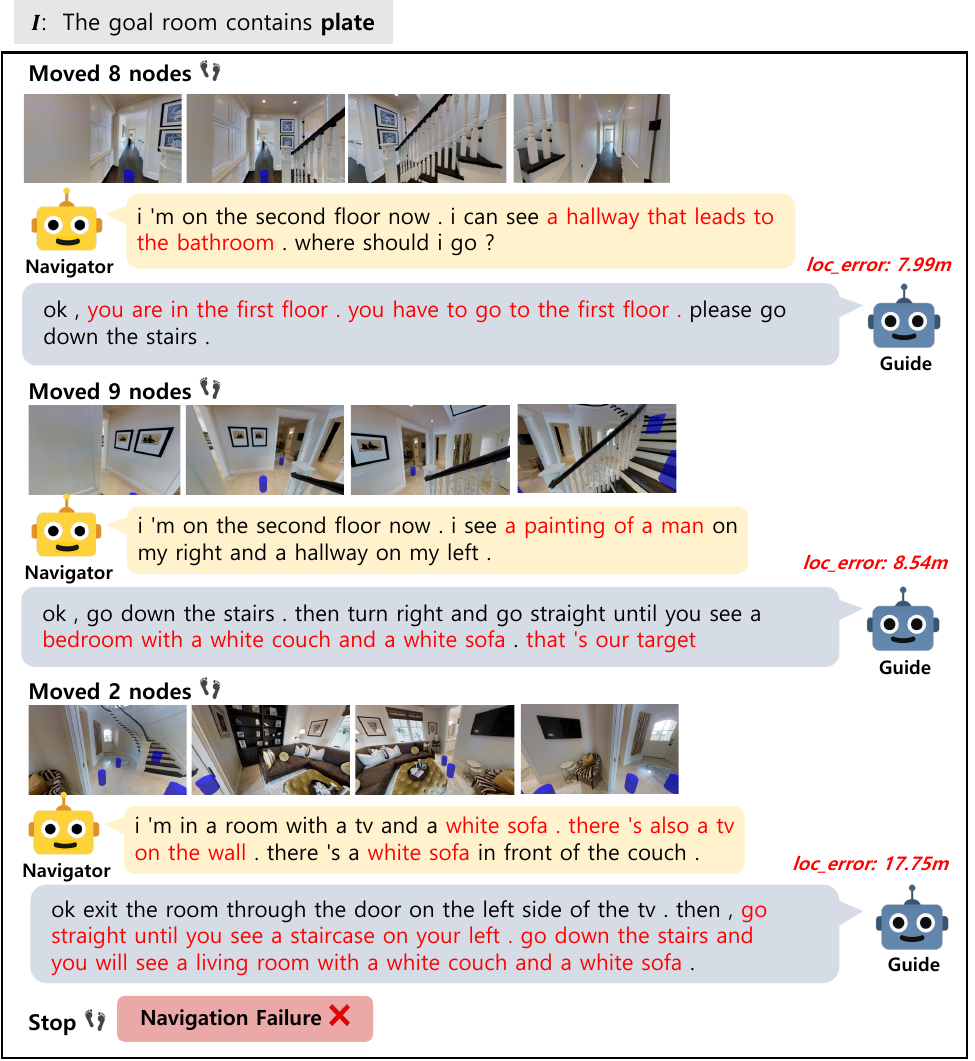} 
        \caption{Baseline~\cite{han2025dialnav}}
    \end{subfigure}
    \hfill 
    \begin{subfigure}[b]{0.49\linewidth}
        \includegraphics[width=\linewidth]{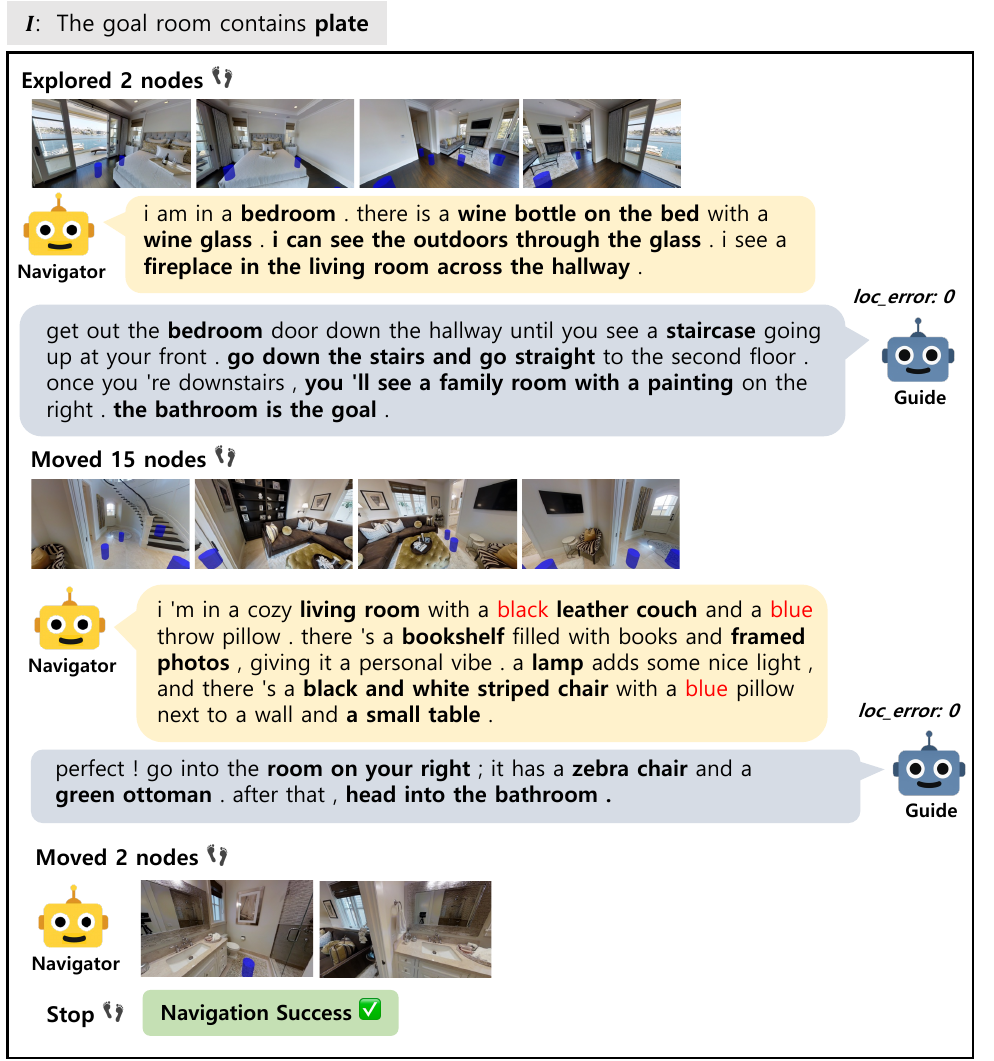}
        \caption{Ours}
    \end{subfigure}

    \caption{
    \textbf{Qualitative comparison on the same task instance between the baseline~\citep{han2025dialnav} (left) and Ours (right).}
    The baseline agent produces broken language with wrong details (marked in \textcolor{red}{red}), likely due to dataset scarcity,
    leading to high localization errors and navigation failure.
    In contrast, our agent provides richer, well-grounded descriptions (marked in \textbf{bold}), yielding accurate localization and reliable instructions, ultimately leading to successful navigation.}
    \label{fig:qualitative_main}
\end{figure*}

\noindent \textbf{Ablations of Localization Model.} \
Table~\ref{tab:localization-ablation} ablates the Graph-based Transformer Localization model. 
To isolate the localization performance, we evaluate it outside the holistic dialog-navigation loop. 
Instead, we use the human-annotated questions from the validation set (the human-human dialog for validation episodes in RAIN); the model is tasked to predict the node from which that question was asked.
We report Localization Error (LE) in meters and 3m accuracy (A@3m), the percentage of predictions within 3m of the ground truth.
Compared to the GCN baseline from prior work~\cite{han2025dialnav} (Row~1), changing the architecture to Graph-based Transformer Localization alone (Row~2) degrades performance, highlighting the difficulty of training this complex model from scratch on scarce data. 
However, initializing it with VLN pretrained weights (Row~3) yields a dramatic improvement, substantially surpassing the baseline (Row~1) in both environments.
This confirms the efficacy of transferring VLN knowledge for localization.
These results demonstrate that GTL, enhanced by knowledge transfer and the RAINbow dataset, effectively improves localization.

\noindent \textbf{Qualitative Examples.} \ \
Figure~\ref{fig:qualitative_main} and Supp. Mat. Sec.~\ref{sec:qualitative} provide qualitative comparisons between the baseline Navigator--Guide agent pair~\citep{han2025dialnav} and ours.

\noindent \textbf{Human-Agent Cooperation.} \ \
To further validate the effectiveness of our approach, we evaluate it in a human--agent cooperation setup (Table~\ref{tab:human-eval-final}).
We build a simulator in which a human participant collaborates with an agent: one takes the role of the Guide, while the other acts as the Navigator.
Our method consistently outperforms the baseline across all roles and environments, demonstrating that it substantially improves human--agent collaboration. Details are provided in Supp. Mat. Sec~\ref{sec:human-agent-cooperation-detail}.

\section{Conclusion}
In this work, we addressed the key limitations of DialNav, a holistic dialog-based navigation task whose performance has been constrained by critical data scarcity. 
We first proposed an automatic data generation pipeline to create RAINbow, a large-scale (238K) dataset that expands the available training data by utilizing existing VLN instructions, addressing the core data scarcity issue. 
Second, we adopted the Dual-Strategy Training scheme for the navigator, aligning the training process with the dynamic, multi-turn nature of the task by combining data-guided rollout and on-policy rollout.
Finally, we adapted a graph-based transformer for the localization, enabling us to leverage powerful VLN pretraining for localization task. 
Our experiments demonstrated that these components are highly synergistic. 
Integrating these solutions, our final model establishes a new state-of-the-art, more than doubling the baseline success rate on both \textit{Val Seen} (30.77 to 58.24, +89\%) and \textit{Val Unseen} (14.52 to 29.05, +100\%) splits. By providing a large-scale, high-quality dataset and adopting training and model to leverage it, this work lays a stronger foundation for future research into more complex, interactive embodied agents.


\section*{Limitations}
While our work substantially improves DialNav through large-scale automatic data generation and training enhancements, several limitations remain. Although RAINbow significantly increases data scale, it is built upon existing VLN trajectories, which may limit environmental and behavioral diversity. In addition, the generated dialogs, while generally natural and goal-consistent, may not fully capture the full richness of human embodied communication. Our evaluation is primarily conducted in indoor navigation settings, and further validation in broader embodied domains would be beneficial. 


\bibliography{custom}

\appendix

\clearpage
\setcounter{page}{1}
\appendix

\section{RAINbow Dataset}
\label{supp:dataset_stats}

\subsection{Qualitative Examples}
\begin{figure*}[t]
    \centering
    \begin{subfigure}{1.0\linewidth}
        \centering
        \includegraphics[width=0.8\linewidth]{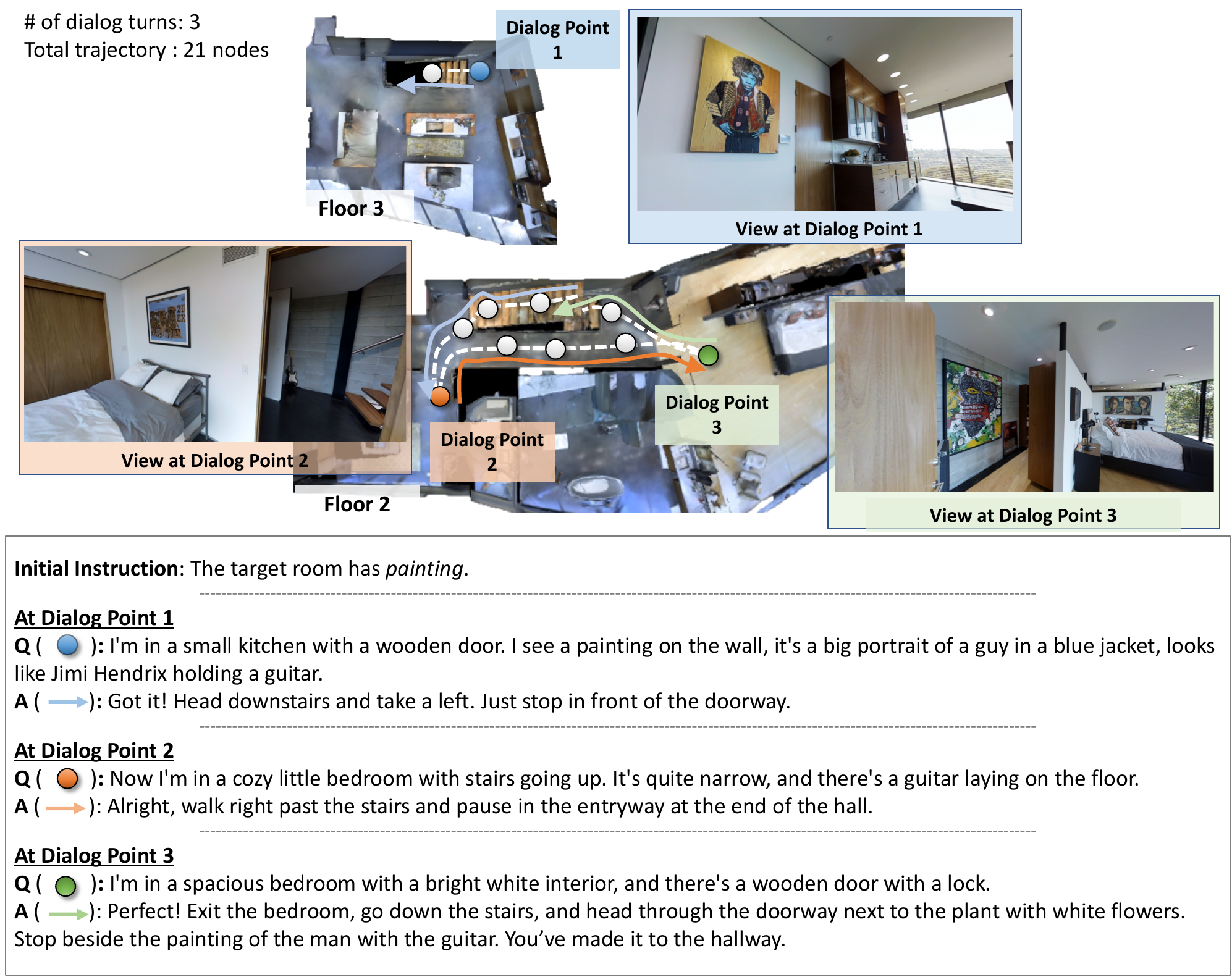}
        \caption{}
    \end{subfigure}
    \begin{subfigure}{1.0\linewidth}
        \centering
        \includegraphics[width=0.8\linewidth]{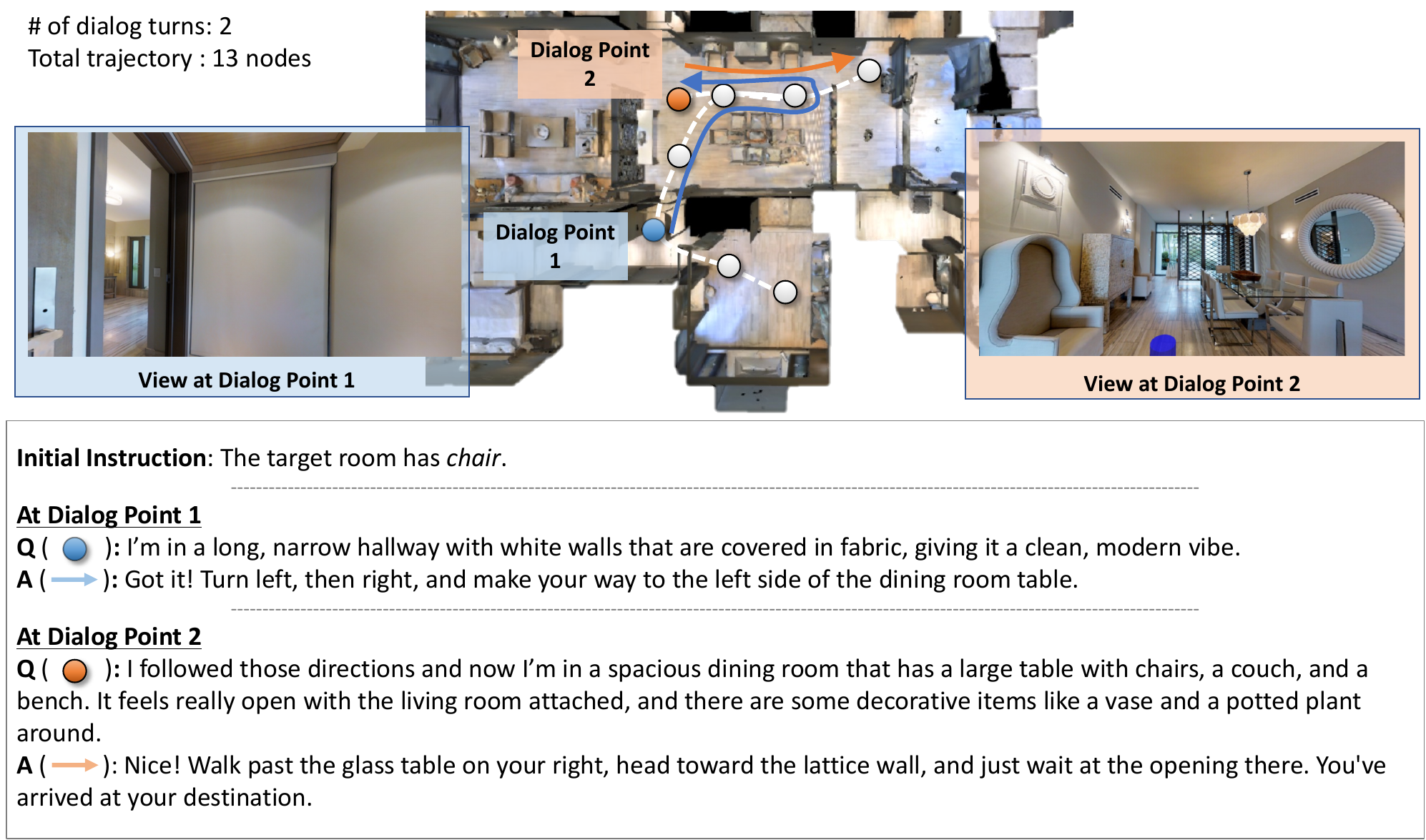}
        \caption{}
    \end{subfigure}
    
    \caption{Examples of RAINbow data (1)}
    \label{fig:rainbow-sample-1}
\end{figure*}

\begin{figure*}[t]
    \centering
    \begin{subfigure}{1.0\linewidth}
        \centering
        \includegraphics[width=0.8\linewidth]{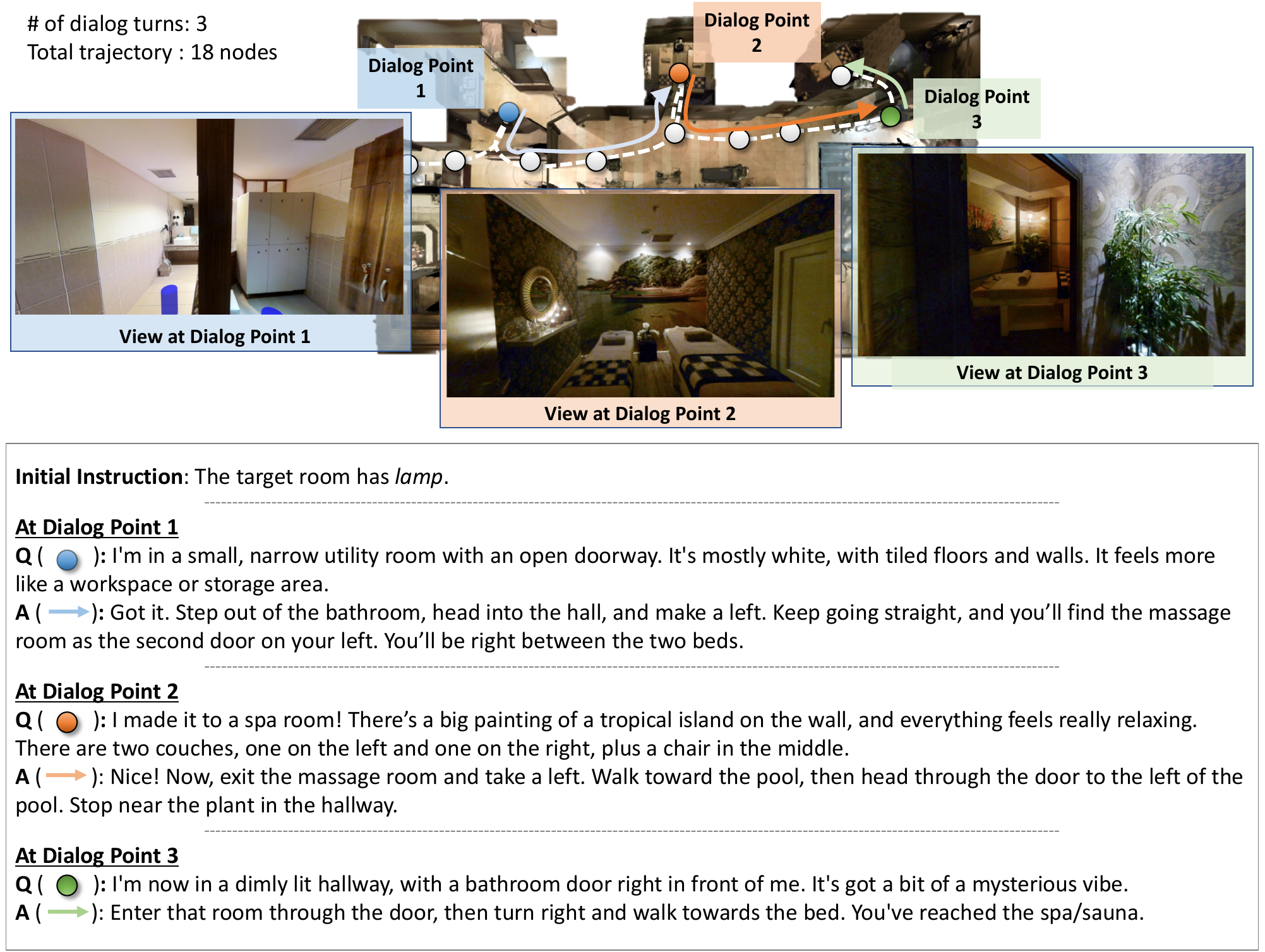}
        \caption{}
    \end{subfigure}
    \begin{subfigure}{1.0\linewidth}
        \centering
        \includegraphics[width=0.8\linewidth]{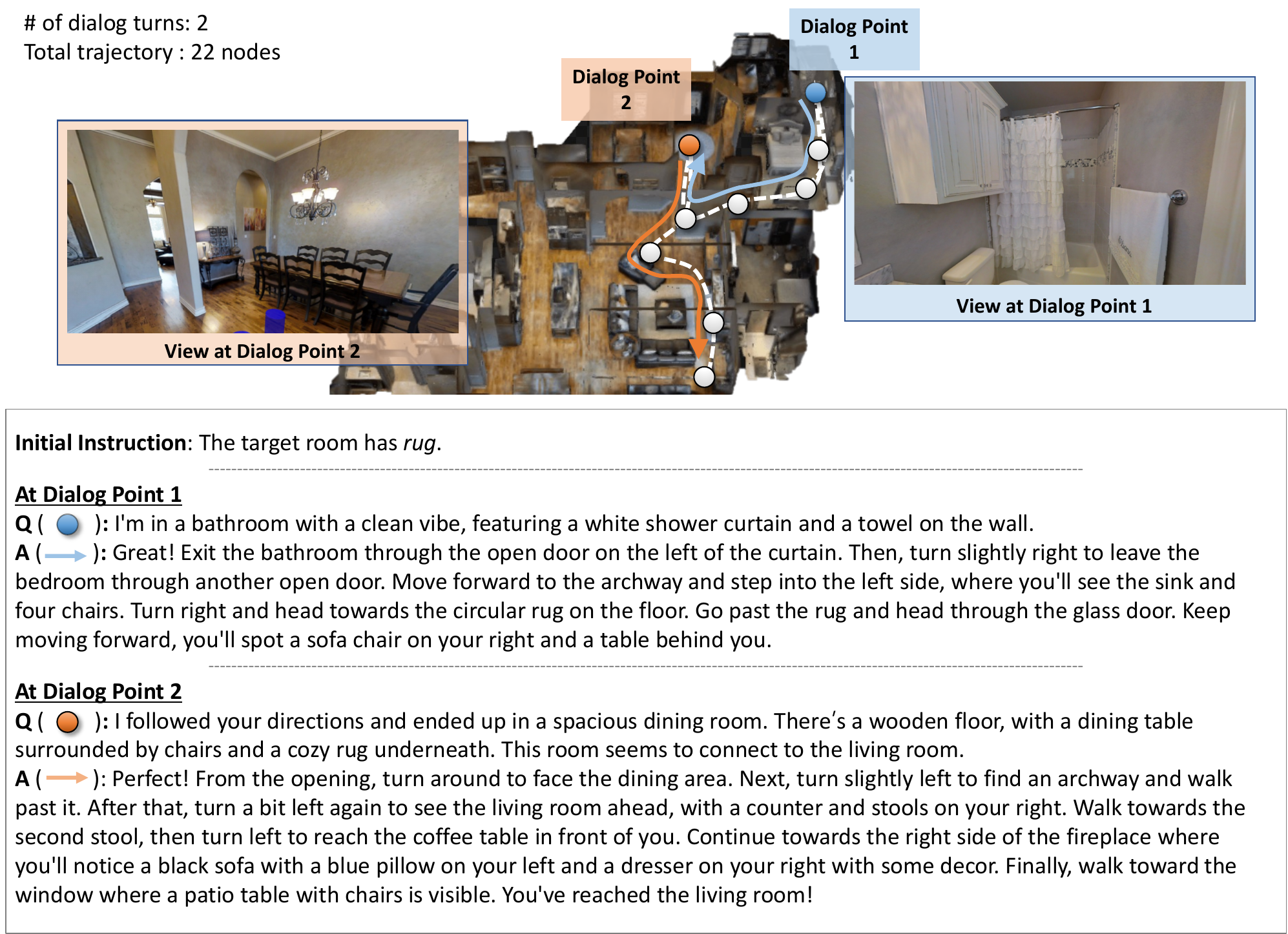}
        \caption{}
    \end{subfigure}
    
    \caption{Examples of RAINbow data (2)}
    \label{fig:rainbow-sample-2}
\end{figure*}

We present qualitative examples from the RAINbow dataset in Figures~\ref{fig:rainbow-sample-1} and~\ref{fig:rainbow-sample-2}.
The dialogs in RAINbow are \textbf{well-grounded} in the environment and the interactions exhibit a \textbf{natural flow}.
The multi-turn exchanges maintain contextual coherence and conversational fluency, closely mimicking the dynamics of human-to-human collaboration.

\subsection{RAIN vs. RAINbow}

We compare the original \textbf{RAIN} dataset (2,233 episodes) with our automatically generated \textbf{Rainbow} dataset (238,028 episodes). 
Table~\ref{tab:dataset-comparison} reports mean, median, maximum, and minimum values across episodes for key statistics.
Figure~\ref{fig:rain-sample} compares dialog style between human annotated RAIN dataset and generated RAINbow dataset.
\begin{table}[ht]
\centering
\resizebox{\linewidth}{!}{
    \begin{tabular}{l ccc ccc}
    \toprule
    & \multicolumn{3}{c}{\textbf{RAIN}} & \multicolumn{3}{c}{\textbf{Rainbow (Ours)}} \\
    \cmidrule(lr){2-4} \cmidrule(lr){5-7}
    \textbf{Statistic} & Mean & Min & Max & Mean & Min & Max \\
    \midrule
    \# Episodes & \multicolumn{3}{c}{2,231} & \multicolumn{3}{c}{238,028} \\
    %
    Dialog turns / ep. & 1.88 & 0 & 8 & 2.71 & 1 & 4 \\
    Question length (words) & 27.48 & 1 & 119 & 30.43 & 4 & 125 \\
    Answer length (words) & 42.53 & 1 & 184 & 43.17 & 0 & 300 \\
    Trajectory length (nodes) & 26.03 & 4 & 116 & 19.54 & 2 & 67 \\
    \bottomrule
    \end{tabular}
}
\caption{\textbf{Comparison of dataset statistics between RAIN and Rainbow.} Rainbow is more than two orders of magnitude larger, with richer dialogs and more diverse language.}
\label{tab:dataset-comparison}
\end{table}

\begin{figure}[h]
    \centering
    \footnotesize

    \begin{subfigure}{\linewidth}
        \centering
        \begin{tabular}{|c p{0.85\linewidth}|}
            \hline
            \rule{0pt}{2.6ex} Q: & Hi. I'm bedroom, which has one bed. On the bed there is two green pillows. And also, there is a big picture, and I can't guess what is it about. It's very simple room. \\
            \rule{0pt}{2.6ex} A: & there are 7 bedrooms in this house. I think you're on the 1st floor. If I'm correct, you'll see a very long corridor outside of the room. Can you see? \\
            \rule{0pt}{2.6ex} Q: & Yes, I can see very long hallway, and also there is a chair in the middle of the hallway.
 \\
            \rule{0pt}{2.6ex} A: & okay, at the end of the corridor, you'll find a staircase on your right. Go all the way up. I mean, there'll be another half story when you think you're at the top, so look carefully :)  At the real end of the staircase, you'll find a kitchen. That's your goal room. \\
            \rule{0pt}{2.6ex} Q: & I think I'm at that kitchen, there is a statue of woman in front of the long wooden table. Am I at the goal room? \\
            \rule{0pt}{2.6ex} A: & Yes correct. good work
 \rule[-0.9ex]{0pt}{0pt} \\
            \hline
        \end{tabular}
        \caption{RAIN}
    \end{subfigure}
    
    \vspace{0.2cm}
    
    \begin{subfigure}{\linewidth}
        \centering
        \begin{tabular}{|c p{0.85\linewidth}|}
            \hline
            \rule{0pt}{2.6ex} Q: & I'm in a small kitchen with a wooden door. I see a painting on the wall, it's a big portrait of a guy in a blue jacket, looks like Jimi Hendrix holding a guitar. \\
            \rule{0pt}{2.6ex} A: &Got it! Head downstairs and take a left. Just stop in front of the doorway. \\
            \rule{0pt}{2.6ex} Q: &Now I'm in a cozy little bedroom with stairs going up. It's quite narrow, and there's a guitar laying on the floor. \\
            \rule{0pt}{2.6ex} A: & Alright, walk right past the stairs and pause in the entryway at the end of the hall.\\
            \rule{0pt}{2.6ex} Q: & I'm in a spacious bedroom with a bright white interior, and there's a wooden door with a lock. \\
            \rule{0pt}{2.6ex} A: & Perfect! Exit the bedroom, go down the stairs, and head through the doorway next to the plant with white flowers. Stop beside the painting of the man with the guitar. You’ve made it to the hallway.
 \rule[-0.9ex]{0pt}{0pt} \\
            \hline
        \end{tabular}
        \caption{RAINbow}
    \end{subfigure}

    \vspace{-0.1cm}
    \caption{\textbf{Qualitative comparison of RAIN (Human) and RAINbow (Generated) dialogs.} The figure presents examples of complex, multi-turn dialogs. The RAIN example (top) illustrates the necessity of detailed scene descriptions for successful remote localization. The RAINbow example (bottom) demonstrates that our automatic generation pipeline successfully replicates this essential dialog structure.}
    \label{fig:rain-sample}
    \vspace{-0.3cm}
\end{figure}


\subsection{CVDN vs. RAINbow}
We provide a qualitative comparison with CVDN. 
Their interaction dynamics differ due to task setups: CVDN (Guide knows Navigator’s position) vs. DialNav (Guide must infer Navigator's location). 
In CVDN, the Navigator's questions often serve as a mere trigger for the next instruction (\eg ``hello", ``where now?"), and each QA turn tends to resemble a sequence of independent \textit{instruction-following} instances. 
Conversely, DialNav requires the Navigator to share context for localization. 
This involves potential misalignments, such as mislocalization or misnavigation, which RAINbow captures in its pipeline (details in Supp. Sec. C.1.7) exhibiting \textit{decision-driven} actions, such as re-querying to resolve uncertainty.
\begin{table}[h]
    \centering
    \vspace{-1em}
    \scriptsize 
    \renewcommand{\arraystretch}{0.8}
    \begin{tabular}{|p{0.48\linewidth}|p{0.45\linewidth}|} 
        \toprule
        \centering \textbf{RAINbow} & \centering \textbf{CVDN} \tabularnewline
        \textit{N:} I'm in a bathroom with green tile walls and a painting over the ... \newline
        \textit{G:} Got it! From the washroom, turn a bit right towards the drawers and then ... \newline
        \textit{N:} \textit{I think I might have gone off track a little.} Right now, I'm in a bright ... \newline
        \textit{G:} No problem, thanks for letting me know. Head out the open door next ... 
        &
        \textit{N:} hello \newline
        \textit{G:} hi, you need to leave the room and go to the room that has four two-seat sofas in it. \newline
        \textit{N:} where now? \newline
        \textit{G:} go up the stairs located beside the table in the room you are at now.
        \tabularnewline
        \bottomrule
    \end{tabular}
    \vspace{-2em}
    \label{tab:qual_comp}
\end{table}
\vspace{0.2em}

\section{Comparison of Embodied Dialog Dataset}

Table~\ref{tab:dataset_comparison} presents a comparison between RAINbow and existing embodied dialog datasets.
Prior datasets typically rely on direct human annotation, which creates a trade-off between quality and cost, resulting in limited scale.
In contrast, RAINbow is constructed by repurposing existing high-quality, human-annotated VLN datasets.
By employing a simple yet effective pipeline to reformat these resources into dialog, we achieve a massive scale-up at a negligible cost per episode.

\begin{table*}[ht]
    \centering
    \scalebox{0.75}{
        \begin{tabular}{l c r r r c}
        \toprule
        \textbf{Dataset} & \textbf{Task} & \textbf{Size} & \textbf{Total Cost} & \textbf{Cost/Ep} & \textbf{Source} \\
        \midrule
        CerealBar~\citep{suhr2019executing} & Game & 1K & \$5.8K & \$5.80 & Human \\
        IGLU~\citep{kiseleva2022interactive} & Game  & 8.9K & - & - & Human \\
        HoloAssist~\citep{wang2023holoassist} & Manipulation  & 2K & - & - & Human \\
        TEACh~\citep{padmakumar2022teach} & Manipulation & 4K & \$105K & \$26.25 & Human \\
        DialFRED~\citep{gao2022dialfred} & Manipulation & 53K & \$10K & \$0.19 & Hybrid \\
        RobotSlang~\citep{banerjee2021robotslang} & Navigation & 0.2K & - & - & Human \\
        TalkTheWalk~\citep{devries2018talkwalknavigatingnew} & Navigation & 10K & - & - & Human \\
        AVDN~\citep{fan-etal-2023-aerial} & Navigation & 3K & - & - & Human \\
        CVDN~\citep{thomason2020vision} & Navigation & 2K & \$7K & \$3.50 & Human \\
        RAIN~\citep{han2025dialnav} & Navigation & 2K & \$5K & \$2.50 & Human \\

        \midrule
        \textbf{RAINbow (Ours)} & \textbf{Navigation} & \textbf{238K} & \textbf{\$0.2K} & \textbf{$<$ \$0.01} & \textbf{Auto} \\
        \bottomrule
        \end{tabular}
    }
    \caption{\textbf{Comparison with existing embodied dialog datasets.} Most existing datasets relying on human annotation are costly, limited in scale, or restricted to synthetic environments. \textbf{Source}: `Human' denotes human-annotated datasets, while `Auto' denotes automatically generated ones.}
    \label{tab:dataset_comparison}
\end{table*}

\section{Data Augmentation Details}
In this section, we describe our scheme to generate realistic, visually well-grounded, and natural-flowing dialog in detail. 
Our pipeline is designed around three stages: (1) trajectory concatenation, (2) detailed caption generation for each node, and (3) reformatting into a natural dialog flow. 
This process ensures the resulting dialog is contextually grounded while preserving the fine-grained details from the well-formed captions and original VLN descriptions.

\subsection{Trajectory Concatenation Details}
\label{supp:traj_concat}

For trajectory concatenation, we applied the following constraints to ensure natural continuity:  
\begin{enumerate}
    \item We concatenated 2-4 trajectories from R2R~\cite{anderson2018vision}, RxR~\cite{ku2020room}, CVDN~\cite{thomason2020vision} answer trajectories into a single episode.  For CVDN dataset, we use its answer and pair its path with the next 5 shortest path from the question node based on its data collecton description.
    \item The endpoint of one trajectory and the start of the next must be within 1 meter in the navigation graph.  
    \item To prevent overly circuitous paths, the detour ratio, the concatenated path length divided by the shortest path length between the start and end nodes, was constrained to be less than 1.3.  
    \item Episodes in which the goal region contained no selectable object were discarded.  
    \item The ambiguous instruction $I$ was derived from the Matterport3D~\cite{chang2017matterport3d} metadata by randomly selecting one visible object in the goal region. To avoid overly generic references, we excluded a predefined set of objects (e.g., \texttt{wall}, \texttt{floor}, \texttt{ceiling}, etc.).
    \item Since goal regions in DialNav correspond to rooms rather than single nodes, we excluded cases where the agent had already reached the goal room before subsequent dialog turns were appended, avoiding unnatural ``post-goal'' interactions.
    \item To further increase diversity, we additionally introduced variations in 10\% of the constructed episodes, simulating potential deviations in real dialog navigation. We consider three types of variations: \emph{mislocalization}, \emph{misnavigation}, and \emph{exploration}. In the case of mislocalization, the Guide intentionally provides an incorrect path description that does not match the Navigator’s true position; the Navigator then proceeds by moving 1--2 nodes randomly from the original location rather than following the erroneous instruction. For misnavigation, the Navigator deviates from the instructed path and follows a randomly chosen alternative route. In exploration, the Navigator continues beyond the instructed trajectory by taking an additional 2--5 random steps after completing the suggested path. By incorporating these cases, we were able to include instructions that were previously filtered out by the strict trajectory-connection criteria, thereby maximizing the utilization of existing VLN data for dialog augmentation.
\end{enumerate}

\subsection{Caption Generation Details} \label{supp:question_generation} 
\begin{figure*}[t]
\centering
\small
\setlength{\tabcolsep}{6pt}
\renewcommand{\arraystretch}{1.25}

\includegraphics[width=0.9\textwidth]{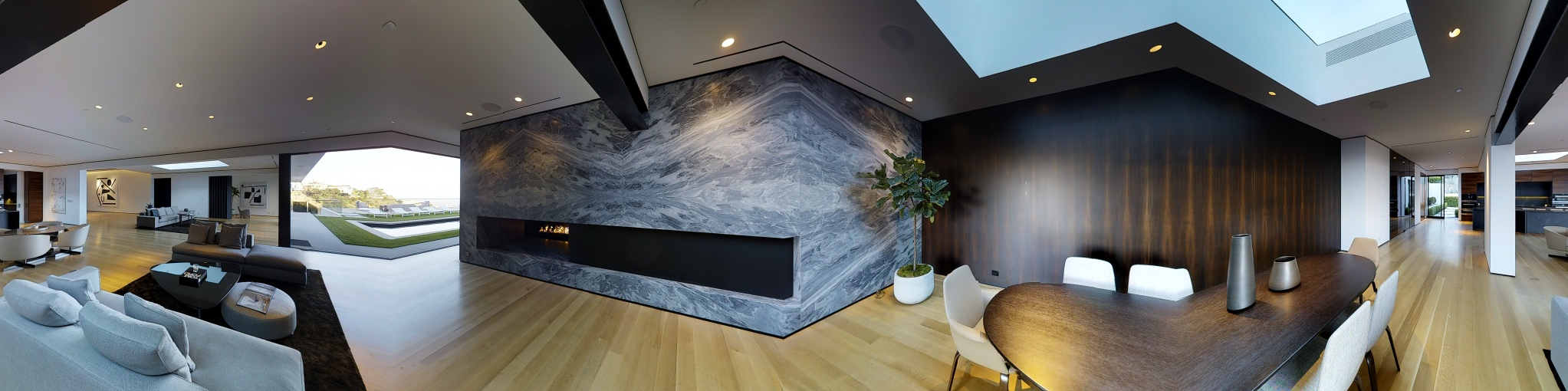}

\begin{tabularx}{\textwidth}{@{}l X@{}}
\toprule
\textbf{Simple caption} & 
\emph{``I'm in a room with a black counter and a large marble wall.''} \\
\midrule

\textbf{Region grounded caption} & 
\emph{``The dining room in this image features a large, modern design with a black and white color scheme. The room is adorned with a marble wall, which adds a touch of elegance to the space. There is a large window that allows natural light to fill the room, creating a bright and inviting atmosphere. A potted plant is placed in the room, adding a touch of greenery and life to the space. Additionally, there is a chair in the room, providing a comfortable seating option for those who wish to dine or relax in the dining area''} \\
\midrule

\textbf{Object caption} & \emph{The dining table in this image is a black table with a vase on it., The pillow is white., The couch in this image is a large, gray sectional sofa., The chair in the image is a modern, sleek design with a gray color. It is positioned at a black dining table., The chair in the image is a modern, minimalist design with a white seat and black legs. It is positioned at a black dining table.} \\
\midrule

\textbf{Region and object grounded caption} & \\
\quad 1-shot, 50\% objects &
\emph{``I'm in a modern dining room with a black and white color scheme, surrounded by a marble wall. There's a large window letting in plenty of natural light, and a potted plant adding some greenery to the space.''} \\
\quad 1-shot, 70\% objects &
\emph{``I'm standing in a modern dining room with a black and white color scheme, surrounded by a marble wall. The room is filled with natural light from a large window and features a potted plant and a chair for seating..''} \\
\quad 3-shot, 50\% objects &
\emph{``I'm in a modern dining room with a black and white color scheme, surrounded by a marble wall. There's a large window letting in plenty of natural light, and a potted plant adds a touch of greenery to the space.''} \\
\quad 3-shot, 70\% objects &
\emph{``I'm in a modern dining room with a black and white color scheme, featuring a marble wall and a large window that lets in plenty of natural light. There's a comfortable chair and a potted plant adding a touch of greenery to the space.''} \\
\quad 5-shot, 50\% objects &
\emph{``I'm standing in a modern dining room with a black and white color scheme, surrounded by a marble wall and a large window that lets in plenty of natural light. A potted plant adds a touch of greenery to the space, and a sleek gray chair sits at a black dining table}. \\
\quad 5-shot, 70\% objects &
\emph{``I'm in a modern dining room with a black and white color scheme. The room features a marble wall, a large window, and a potted plant, creating a bright and inviting atmosphere. There's a chair at a black dining table, providing a comfortable seating option.''} \\
\bottomrule
\end{tabularx}

\caption{Example of generated captions from three different captioning schemes. }
\label{fig:qgen_examples_multi}
\end{figure*}

To produce natural and visually grounded questions, we prompted LLaVA-1.5-7B~\cite{liu2023llava} for visual description and Llama-3.1-8B for text synthesis  with different visual/textual contexts. 
Our aim was to ensure that the resulting utterances contained sufficient local detail for Guide-side localization, while generating diverse content even for the same node. 
We experimented with three variants and randomly used one of these schemes for caption generation for each episode. 
The prompt for each variant is provided in Figure~\ref{fig:prompt_q_panoramic} and the resulting captions are provided in Figure~\ref{fig:qgen_examples_multi}.

\textbf{(A) Simple panoramic caption}: The panoramic observation at a node is directly provided to LLaVA with a short instruction to generate a navigation-oriented question. This produces concise questions.

\textbf{(B) Region-grounded caption}: Here, we first get the room type of each node from Matterpot3D metadata. Then, we instructed LLaVA to provide a description of the panoramic scene, explicitly mentioning the room type. The room types are provided in Matterport3D metadata paired to each navigation node. This yields richer, more descriptive questions.

\textbf{(C) Region and object-grounded panoramic caption}: For a more detailed and visually grounded caption, we first generate captions for visible objects in the given node. Based on Matterport3D metadata, we first extract image patches of all visible objects at a given node. Then, we prompt LLaVA to generate a description of the given object image. To avoid hallucination or errors from wrong metadata annotations, we prompt it to generate 'none' if the given object is not visible in the image. We then jointly supply three inputs to LLaMA-3.1-8B: a general panoramic caption and a randomly sampled subset of the valid object-level captions. LLaMA is then prompted to generate a single detailed caption that integrates both global layout and salient object cues. To increase diversity, we vary the sampling ratio and the number of in-context examples. For in-context samples, we find similar samples in the RAIN dataset based on room types and visible objects.

\subsection{Dialog Reformat Details} \label{supp:reformat} 
The primitive Q\&A content, derived from raw captions and VLN instructions, is less natural than human conversation. To reformat this content into a natural multi-turn dialog, we employ GPT-4o-mini. To ensure formatting consistency and a natural flow, we perform this in two steps. The prompts for these steps are provided in Figure~\ref{fig:prompt_dialog_reformat}.
The result of each steps are provided in Figure~\ref{fig:dialog_reformat_combined}. 
Through this 2 step reformatting, the resulting dialog is natural, visually well grounded and better reflects realistic Navigator-Guide interactions.

\begin{enumerate} 
\item \textbf{Instruction Refining.} First, we refine the raw VLN instructions. We prompt the LLM to make each instruction concise while preserving key spatial references, preventing overly verbose text that is unnatural in conversation. Moreover, since the original VLN instructions are for single-turn tasks, they often prematurely mention reaching the final goal. To prevent this, we also instruct the LLM to generate two versions of each refined instruction: a `goal-oriented' variant (e.g., "You have reached your destination") and a `neutral' variant (which does not mention the goal). We use the goal-oriented variant only for the final dialog turn and the neutral version for all intermediate turns. This explicit branching strategy prevents ambiguous or inconsistent goal mentions.

\item \textbf{Conversational Smoothing.} We then construct a draft dialog by sequencing the original scene captions and their corresponding refined, goal-conditioned instructions (from Step 1). This entire sequence is then paraphrased by GPT-4o-mini into fluent, conversational language, resulting in the final multi-turn dialog. This step retains all navigation-critical details while simplifying overly formal phrases, removing redundancies, and inserting subtle acknowledgments where appropriate to enhance conversational flow. 
\end{enumerate}

\subsection{Dataset Evaluation.}  \label{supp:dataset_evaluation} 
To validate our final dataset we conducted a human evaluation. 30 samples each from RAIN and RAINbow were randomly shuffled and evaluated by 6 annotators.
\textbf{(1) Goal Alignment:} When verifying if the dialog remains faithful to the initial goal, RAINbow achieved 90.0\% accuracy (comparable to RAIN's 93.3\%). This confirms that our \textit{Instruction Refining} step (Supp. Sec. C.3.1) successfully preserves goal consistency.
\textbf{(2) Naturalness:}  In terms of linguistic flow, RAINbow achieved a score of 4.76/5.0, closely mirroring the human-written RAIN baseline 4.83/5.0.

\begin{table}[t]
\centering
    \scalebox{0.9}{
\begin{tabular}{lccc}
\toprule
Split & \# Objects & \# Hallucination & Ratio \\
\midrule
Questions &  4.74 & 0.58 & 10.7\% \\
Answers &  4.77 & 0.13 & 2.0\% \\
\bottomrule
\end{tabular}
}
\caption{\textbf{Average number of objects and hallucinations per turn.}}
\label{tab:hallucination}
\end{table}

\subsection{Hallucination Analysis}
\label{supp:hallucination-analysis}
To further assess the reliability of our pipeline, we conduct a hallucination evaluation on a sampled subset of RAINbow. 

\textbf{Evaluation protocol.} We randomly sampled 100 episodes (247 dialog turns in total) and used a closed-source VLM (Claude Opus 4.7) to verify each object mentioned in the dialog against the actual observation. The table~\ref{tab:hallucination} reports the average number of hallucinated objects per-turn. 

\textbf{Results.} Our pipeline directly repurposes the human-annotated VLN instructions as answers, which results in low hallucination ratio (2.0\%). The hallucination rate in questions is 10.7\%. This low hallucination rate confirms the effectiveness of our pipeline.

\section{Results on CVDN}
\begin{table}[ht]
\centering
\resizebox{\linewidth}{!}{
\begin{tabular}{l c c}
\toprule
\textbf{Method} & \textbf{Val Seen (m)} & \textbf{Val Unseen (m)} \\
\midrule
Shortest Path & 32.8 & 29.3 \\
RMM$_{N=3}$ \citep{roman2020rmm} & 14.0 & 5.6 \\
SCoA~\citep{zhu2021self} & 19.5 & 11.2 \\
\midrule
\textbf{Ours} & \textbf{26.5} & \textbf{15.7} \\
\bottomrule
\end{tabular}
}
\caption{\textbf{Evaluation on the CVDN Dialog Setup.}The table compares our method against established baselines on the CVDN dataset \cite{thomason2020vision}. The performance is \textbf{Goal Progress (m)}, where higher values indicate better performance.}
\label{tab:cvdn-goal-progress}
\end{table}

We evaluate our agents on the established Vision-and-Dialog Navigation (CVDN) dataset~\cite{hao2020towards}.
Table \ref{tab:cvdn-goal-progress} presents a comparison against methods evaluated under the dialog in a loop setup. 
The CVDN evaluation uses a simplified remote Guide configuration: the Guide is provided with the Ground Truth localization and limits its guidance to 5 future navigation steps.
Our approach achieves superior performance over all listed methods, demonstrating the strong generalization effect of our large-scale RAINbow dataset and training scheme to this external domain.

\section{Model Details}
\label{sec:model-detail}

\subsection{Agent Capabilities and Information Flow}
The overall DialNav task relies on the specialized capabilities and distinct information access of the two agents. 
The integrated agents (Navigator and Guide) possess the full knowledge boundaries listed below. 
\textbf{Navigator Agent} (Navigation, Question, Whether-to-ask Decision): The Navigator has access to the accumulated dialog history, its navigational history (the subgraph of visited nodes and adjacent locations), and the initial instruction ($I$).
\textbf{Guide Agent} (Localization, Answer): The remote Guide has knowledge of the full environment graph ($G$), the initial instruction ($I$), and the goal region ($R$). Since the Guide knows the full environment $G$, it can calculate the shortest path between any two nodes.
However, i this work, for implementation and training convenience, each submodule utilized only the specific input features that its model architecture was designed to accept.

\subsection{Navigation Model}
\label{sec:navigation-model-detail}
For the core action prediction module (Navigator), we adopt the Dual-scale graph Transformer (\textbf{DUET})~\cite{chen2022think} architecture, following established practices in DialNav research~\cite{han2025dialnav}. DUET is a strong VLN backbone designed for joint long-term action planning and fine-grained cross-modal understanding.

\textbf{Architecture and Reasoning.} DUET explicitly builds a topological map and dynamically combines two scales of encoding via graph transformers: a fine-scale encoding of local observations and a coarse-scale encoding of the global map. This dual-scale approach allows the Navigator to reason efficiently over a large action space.

\textbf{Pretraining.} We leverage publicly available weights from large-scale VLN pretraining frameworks, such as ScaleVLN~\cite{wang2023scaling}. This initialization is vital because the pretraining process exposes the model to millions of diverse instruction-trajectory pairs across various environments, imparting robust environmental knowledge that is crucial for significantly reducing the generalization gap between navigating in seen and unseen environments.

\begin{itemize}
    \item \textbf{Input:} Accumulated visual and spatial history, initial instruction and the latest navigation instruction appended as single instruction.
    \item \textbf{Output:} Probability distribution over the next optimal action $v_{t+1}$ (move or stop).
\end{itemize}

\subsection{Question and Answer Generation Model}
For both question and answer generation, we adopt the \textbf{LANA} model~\cite{wang2023lana} as our baseline module. LANA is a vision-and-language model originally designed to enhance navigation performance by jointly learning to describe past and upcoming paths.

\begin{itemize}
    \item \textbf{Input:} Panoramic observation ($X_v$) for question generation; or a sequence of panoramic images for the next subpath for answer generation.
    \item \textbf{Output:} Natural language question ($q$) or guiding answer ($a$).
\end{itemize}
In our setup, LANA is initialized with its proposed pretraining and fine-tuned on the RAINbow and RAIN datasets to adapt navigation-style instructions into dialog-style questions and answers.

\subsection{Localization Model}

For localization, we use a lightweight cross-modal architecture, the Graph Convolutional Network (GCN) Localization~\cite{hahn2022transformer}, as our reproducible baseline. The GCN model formulates localization as a \emph{node-selection} problem.
We introduce our graph-based localization, adopting a transformer inspired by \textbf{DUET}~\cite{chen2022think} to improve position inference from dialog.

\textbf{Repurposing the VLN Architecture.} We adapt the DUET architecture to leverage knowledge from VLN pretraining. For localization, we repurpose the architecture by substituting the inputs and reinterpreting the output:
\begin{itemize}
    \item \textbf{VLN Input} $\rightarrow$ \textbf{Localization Input:} The navigation instruction $I$ is replaced with the \textbf{Navigator's question $q$}, and the partial navigation map $\hat{G}_t$ is replaced with the \textbf{entire house graph $G$}.
    \item \textbf{VLN Output} $\rightarrow$ \textbf{Localization Output:} The model is tasked to select the most probable node corresponding to the Navigator's current position $v_q$ from all nodes in $G$: $v_q = \text{DUET}_{\text{LOC}}(G, q)$.
\end{itemize}
This adaptation, leveraging large-scale pretrained weights, provides robust perceptual grounding and improves localization accuracy.

\subsection{Whether-to-Ask}
The \textit{whether-to-ask} module governs the interactive nature of the Navigator, deciding when to request external guidance. This module uses a confidence thresholding strategy, where the decision to ask or proceed autonomously is determined by the model's confidence in its immediate next action.
\begin{itemize}
    \item \textbf{Input:} The Navigator's probability distribution over the next possible actions $p(v_{t+1}| \text{context})$.
    \item \textbf{Mechanism:} The most confident action in next possible actions is treated as our candidate action. If the candidate action's probability falls below a fixed threshold $\tau$, the Navigator initiates a question $q_j$.
    \item \textbf{Output:} Binary decision (Ask=1 or Proceed=0).
\end{itemize}

\section{Localization Model on WAY Task}
The \textbf{Where Are You? (WAY)}~\cite{hahn2020you} is a foundational benchmark for embodied localization, requiring a model to predict its current position $v_q$ within a visual environment (Matterport3D) based solely on a textual query. This task is crucial as it isolates the language grounding ability required by the remote Guide setup in DialNav.

Our model, which integrates the effects of large-scale pretraining from ScaleVLN and the data diversity provided by the RAINbow dataset, has achieved strong performance in the holistic DialNav setup.

We further fine-tune our graph-based transformer localization model on the WAY benchmark to demonstrate that our architecture generalizes well to existing localization tasks. 
Table~\ref{tab:way_localization_results} provides the results of our model against established single-shot baselines for the WAY localization task, where our model shows superior performance.

\begin{table}[t]
\centering

\resizebox{\linewidth}{!}{
\begin{tabular}{l c c c c}
\toprule
\textbf{Method} &
\multicolumn{2}{c}{\textbf{Val Seen}} &
\multicolumn{2}{c}{\textbf{Val Unseen}} \\
\cmidrule(lr){2-3} \cmidrule(lr){4-5}
& Acc@0m $\uparrow$ & LE $\downarrow$
& Acc@0m $\uparrow$ & LE $\downarrow$ \\
\midrule
Human & 47.87 & 6.00 & 56.13 & 3.20 \\
\midrule

Random & 0.33 & 20.80 & 1.90 & 18.61 \\
GCN~\cite{hahn2022transformer} & 19.67 & 10.95 & 8.64 & 9.10 \\
LED-Bert~\cite{hahn2022transformer} & 25.57 & 9.04 & 21.07 & 8.82 \\
DiaLoc~\cite{zhang2024dialoc} & 25.64 & - & 7.02 & - \\
CGD~\cite{wang2025towards} & 30.21 & - & 16.72 & - \\
\midrule
\textbf{Ours} & \textbf{32.46} &\textbf{6.85}& \textbf{23.31} & \textbf{7.75} \\
\bottomrule
\end{tabular}
}
\caption{\textbf{Localization Performance on the WAY Dataset (Single-Shot).}
Comparison of our graph-based transformer model against established single-shot baselines.
We report accuracy for exact node prediction (Acc@0m) and Localization Error (LE) in meters across seen and unseen environments.}
\label{tab:way_localization_results}
\end{table}

\section{Further Analysis on Table~\ref{tab:holistic-results}}

\begin{table}[t]
\centering
\scalebox{0.7}{
\begin{tabular}{lccccc}
\toprule
\multirow{2}{*}{\textbf{Model}} & \multirow{2}{*}{\textbf{SR}} 
& \multicolumn{2}{c}{\textbf{NSC}} & \multicolumn{2}{c}{\textbf{DTC}} \\
\cmidrule(lr){3-4} \cmidrule(lr){5-6}
& & \textbf{All} & \textbf{Succ.} & \textbf{All} & \textbf{Succ.} \\
\midrule
\multicolumn{6}{l}{\textit{Val Seen}} \\
Baseline~\cite{han2025dialnav}   & $58.2$ & $21.30$ & $19.66$ & $4.66$ & $3.49$ \\
Ours & $30.8$ & $18.65$ & $19.89$ & $3.44$ & $3.57$ \\
\midrule
\multicolumn{6}{l}{\textit{Val Unseen}} \\
Baseline~\cite{han2025dialnav} & $14.5$ & $15.52$ & $14.29$ & $5.61$ & $5.03$ \\
Ours & $29.0$ & $21.46$ & $20.26$ & $11.08$ & $8.23$ \\
\bottomrule
\end{tabular}
}
\caption{\textbf{Trajectory statistics across success cases.} Comparison of Success Rate (SR), Navigation Step Count (NSC), and Dialog Turn Count(DTC) for all episodes (All) and successful episodes only (Succ.).}
\label{tab:success_failure_stats}
\vspace{-0.3cm}
\end{table}

Overall, SPL, NE, and OSR exhibit trends consistent with SR. Separately, NSC and DTC demonstrate that RAINbow and DST encourage exploration in uncertain trajectories. To verify this, we measured NSC and DTC specifically within successful episodes for both the baseline and the models equipped with RAINbow and DST. In the Val Seen split, both scores remain nearly identical to the baseline (DTC: 3.57 → 3.49; NSC: 19.89 → 19.66), indicating that the overall increases in these metrics stem primarily from exploratory behavior in uncertain cases. Building on this, the Val Unseen results show that this uncertainty-driven exploration leads to improved generalization to unseen environments: the models achieve nearly double the success rate while exhibiting increased exploration even within successful episodes (DTC: 5.03 → 8.23; NSC: 14.29 → 20.26).

\section{Qualitative Results}
\label{sec:qualitative}

\subsection{Holistic Navigation}

\begin{figure*}[t]
    \centering
    \includegraphics[width=0.8\linewidth]{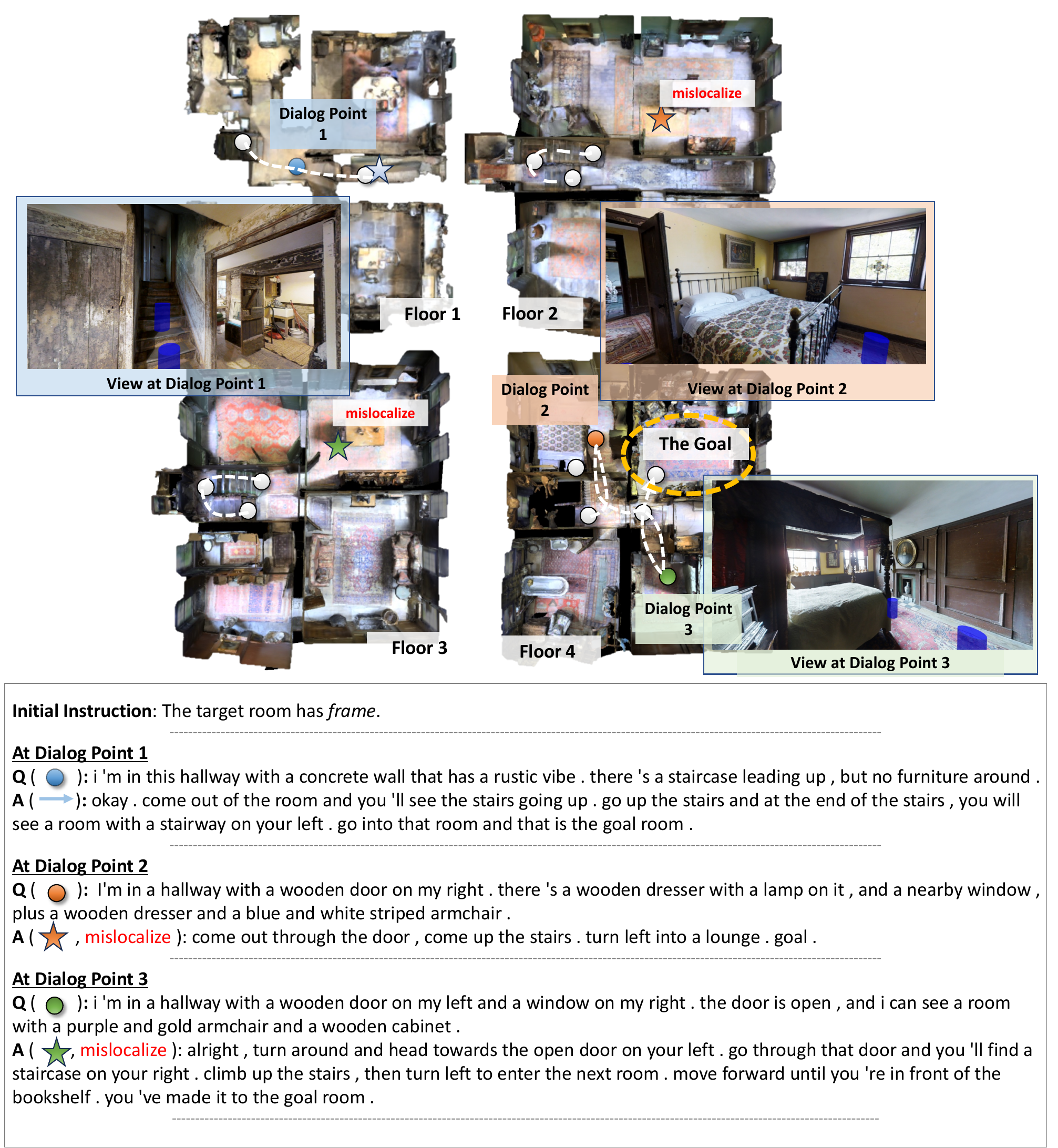}
    \caption{\textbf{Successful example of DialNav episode of our Navigator and Guide agent in Val Unseen environment.} Although the agent failed to localize its current position in the complex environment, it successfully acquired crucial information about the target room through dialogue. This allowed the agent to self-recover its trajectory and successfully navigate to the destination.}
    \label{fig:unseen-sample-2}
\end{figure*}

\textbf{Success with Self-Recovery Despite Localization Failure.} Figure \ref{fig:unseen-sample-2} illustrates a successful DialNav episode that strongly demonstrates the self-recovery capability afforded by our integrated framework. Initially, the Navigator faced a complex environment where the Guide failed to accurately determine the current position. Despite receiving this falsely grounded instruction, the Navigator successfully overrode the erroneous localization estimate and dynamically reconstructed its trajectory (self-recovery) based on goal-relevant information within the dialog.

\begin{figure*}[t]
    \centering
    \includegraphics[width=0.8\linewidth]{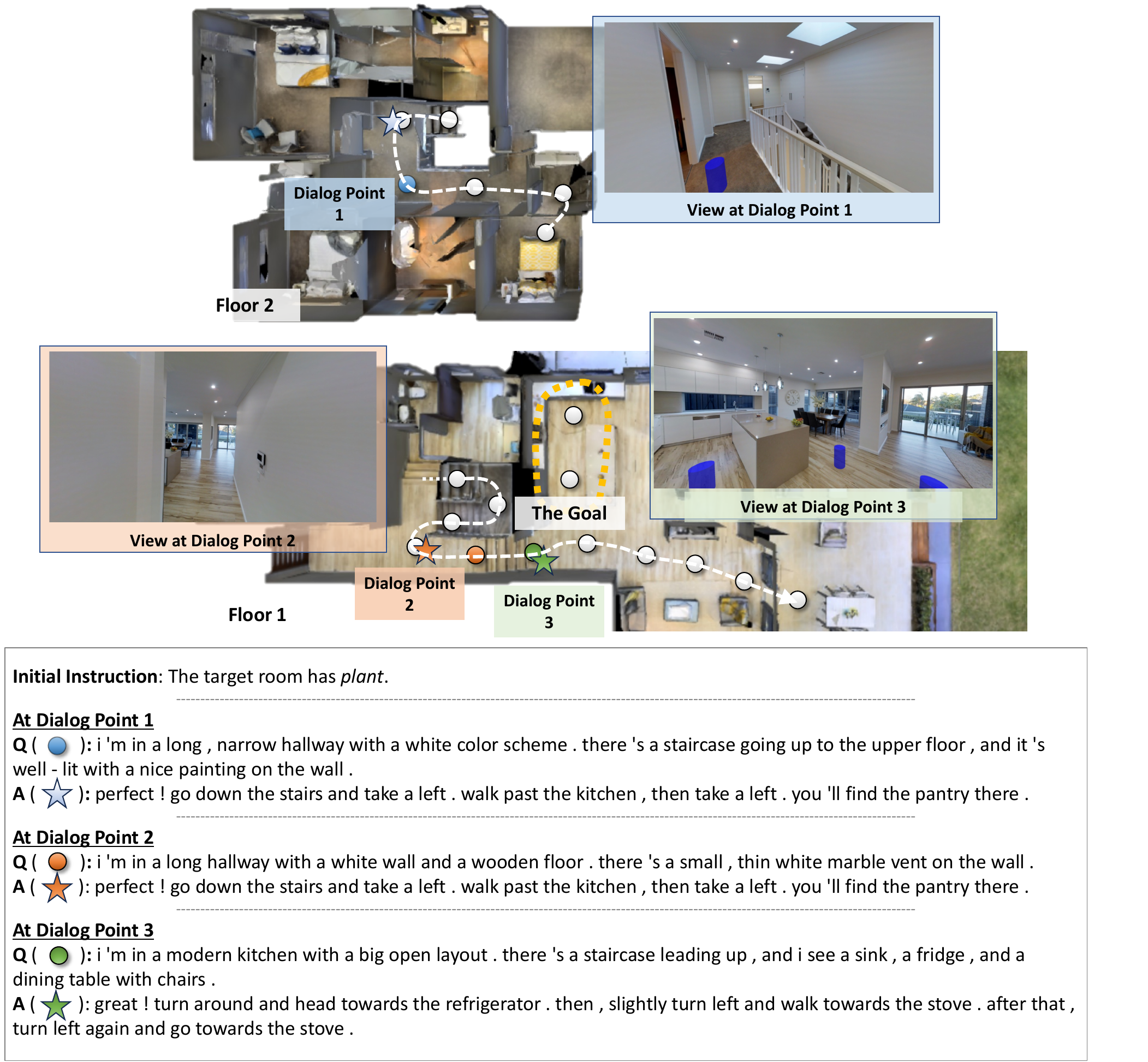}
    \caption{\textbf{Failure case of DialNav episode of our Navigator and Guide agent in Val Unseen environment. }Despite achieving successful localization and receiving a well-grounded answer regarding the path, the navigation agent failed to follow the instruction derived from the dialogue, resulting in a failure to reach the destination.
    }
    \label{fig:unseen-sample-1}
\end{figure*}

\noindent\textbf{Failure Due to Instruction-Execution Inconsistency.} Conversely, Figure \ref{fig:unseen-sample-1} presents a critical failure case that underscores the remaining challenge in seamlessly integrating the conversational policy with the motion execution policy. In this episode, the Guide successfully localized the Navigator and provided a well-grounded answer. However, the Navigator failed to accurately execute the trajectory dictated by the dialogue's instructions, resulting in navigation failure.
\label{sec:qualitative-results}

\subsection{Generated Dialog}

We provide a qualitative comparison between dialogs generated by the baseline (trained only on RAIN) and our enhanced model (trained on RAINbow) against the human-annotated Ground Truth (GT) in Table \ref{tab:qualitative_questions_and_answers}. 
All generated samples were produced under the identical preceding dialog context and current panoramic viewpoint.
This analysis clearly demonstrates the improvement afforded by large-scale data augmentation.

The baseline model trained solely on the small RAIN dataset often suffers from redundancy and fragmentation. For instance, in the Answer Example, the RAIN model enters a circular loop ("you have to go to the dining room") and omits essential path information. Similarly, in the Question Example, the generated question is highly repetitive ("a fireplace, a fireplace, and a fireplace"), failing to provide sufficient, diverse visual cues for the Guide to localize the agent.
In contrast, the model fine-tuned on \textbf{RAINbow} produces significantly more fluent and descriptive instructions.
The RAINbow answers include detailed navigational steps grounded in specific objects ("exit the room and turn right to find a staircase") and generally maintain a coherent conversational flow.
The RAINbow questions successfully utilize rich visual features ("a wooden table right in the middle," "a large, sliding glass door to the outside") to generate contextually relevant descriptions, which is vital for localization.

However, a qualitative gap still exists compared to the human-written Ground Truth. While RAINbow greatly improves fluency, it sometimes lacks the fine-grained object-level details. Nonetheless, the substantial gain in fluency and coherence achieved by RAINbow validates the effectiveness of our automatic generation pipeline for boosting dialog quality.

\section{Analysis of Confidence Threshold}

\begin{figure}[ht]
\centering
\begin{subfigure}{0.75\linewidth}
\centering
\includegraphics[width=\linewidth]{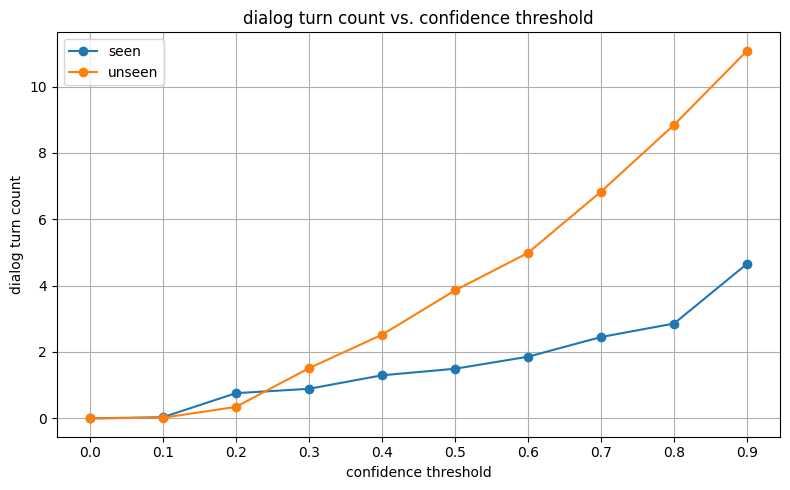} 
\caption{Dialog Turn Count vs. Confidence Threshold}
\label{fig:conf_sr}
\end{subfigure}
\vspace{1em} 
\begin{subfigure}{0.75\linewidth}
\centering
\includegraphics[width=\linewidth]{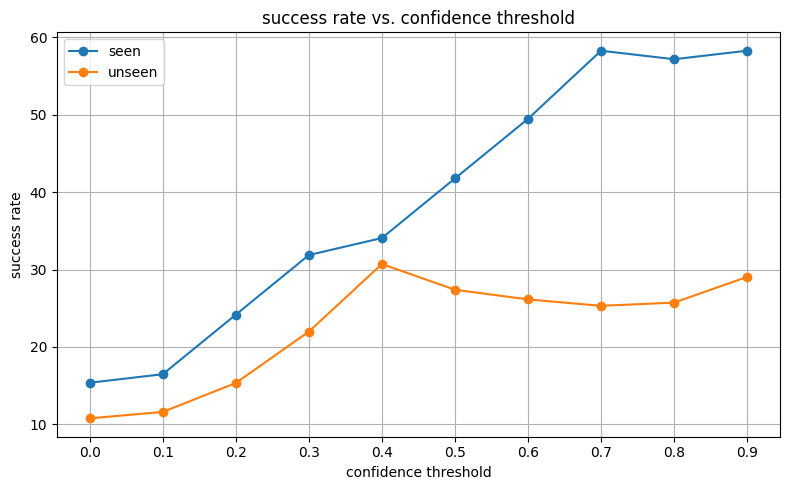} 
\caption{Success Rate vs. Confidence Threshold}
\label{fig:conf_steps}
\end{subfigure}
\vspace{1em} 
\begin{subfigure}{0.75\linewidth}
\centering
\includegraphics[width=\linewidth]{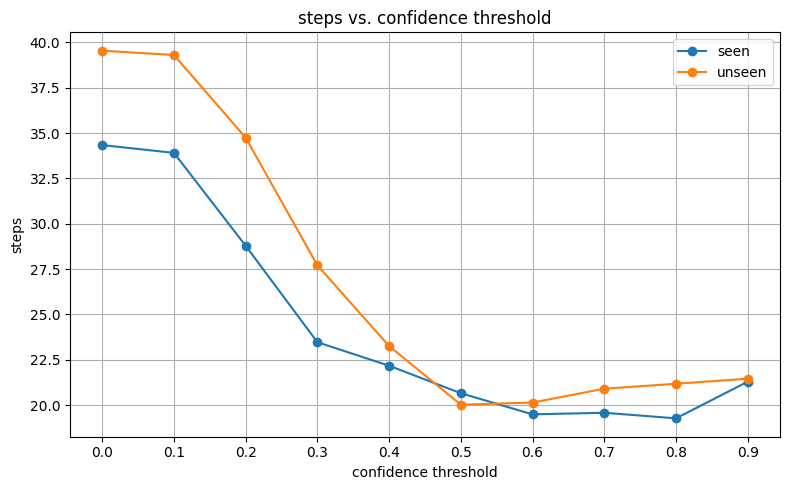} 
\caption{Navigation Steps vs. Confidence Threshold}
\label{fig:conf_dialog}
\end{subfigure}
\caption{\textbf{Impact of Confidence Threshold ($\tau$) on performance.} Adjusting the threshold value allows the Navigator to balance autonomous navigation with dialog requests, showing the trade-off between task success and exploration efficiency.}
\label{fig:confidence_ablation}
\end{figure}

We analyze the impact of the Confidence Threshold ($\tau$), the hyperparameter governing the Navigator's decision to ask for help. As the threshold $\tau$ increases, the agent triggers a question more often when facing uncertainty. This phenomenon is clearly demonstrated by the Dialog Turn Count (DTC), which shows a consistent increase across both splits as $\tau$ approaches $0.9$ (Figure~\ref{fig:conf_sr}).

The increase in dialog frequency directly correlates with the final navigation Success Rate (SR)(Figure~\ref{fig:conf_steps}). In \textit{Val Seen}, SR shows a sustained increase as the threshold rises. This indicates that dialog is effective at resolving uncertainty and guiding the agent to success in familiar environments. In \textit{Val Unseen}, SR increases up to $\tau \approx 0.5$ and then saturates. This contrast suggests that while dialog provides essential initial information, the model reaches a performance ceiling in novel environments, possibly limited by the quality of answer in complex, unseen settings.
The difference in saturation points confirms that the benefit of dynamic dialog is substantially greater in known environments, where the learned policy can best leverage the updated instruction for planning.

Furthermore, we observe that dialog significantly impacts exploration efficiency (Figure~\ref{fig:conf_dialog}). Across both Val Seen and Val Unseen, the Navigation Step Count (NSC) decreases up to $\tau \approx 0.5$ before stabilizing. This demonstrates that proactive dialog is highly effective at reducing redundant exploration, as the agent quickly seeks guidance instead of wandering when uncertain. The stable reduction in NSC followed by stabilization in SR (after $\tau=0.5$ in Val Unseen) shows a clear correlation between reducing navigational uncertainty via dialog and achieving task success.

\section{Details on Human-Agent Cooperation}
\label{sec:human-agent-cooperation-detail}

\textbf{Participants.} Ten participants were recruited internally from our research group. All participants signed consent form consent prior to the evaluation. 

\textbf{Task protocol.} Each participant interacted with our agent through a GUI simulator that supported mouse-based navigation in the Matterport3D environment paired with a text-based dialog window. Participants took the role of either Navigator (sending questions and navigating based on the agent's answers) or Guide (receiving the participant Navigator's questions and providing localization-grounded answers), while the agent took the complementary role. The four conditions in Table 5 — {Navigator, Guide} × {Baseline, Ours} — were presented in randomized order across both Val Seen and Val Unseen splits. Each participant completed 20–30 episodes in total. After each episode, participants rated the agent counterpart's helpfulness on a 1–5 Likert scale, which we report as Human Score (HS) in Table 5; Success Rate (SR) was computed from task outcomes. 

\textbf{Ethics/IRB.} Our human evaluation qualifies for IRB exemption under our institution's policy, as it does not identify individual participants and does not collect or record sensitive personal information as defined by applicable data protection regulations. The evaluation collected only task outcomes (navigation success) and helpfulness ratings; no personal identifiers, demographic information, or sensitive data were recorded.

\section{Potential Risks}
\label{sec:potential-risks}

Our work studies dialog-enabled embodied navigation, where language directly affects action decisions in a physical environment. As such systems become more capable, incorrect or misleading dialog may lead to unsafe or inefficient actions. This is particularly important in embodied settings, where execution errors may carry physical consequences.

In addition, our automatic data generation pipeline uses large language and vision--language models to synthesize dialog. Although we validate faithfulness to the intended goal and manually inspect examples, generated data may still contain occasional inaccuracies, biases, or unnatural interaction patterns. If used without careful validation, such artifacts could propagate into trained embodied agents.

We mitigate these risks in three ways. First, our task is evaluated in simulation rather than physical deployment. Second, we validate generated dialog quality and goal faithfulness through both automatic design constraints and human evaluation. Future work should further investigate robustness to adversarial, ambiguous, or misleading dialog in real-world embodied settings.

\section{Data Safety, Privacy, and Content Checks}
\label{sec:data-safety}

We discuss here whether the data used or created in this work contains personally identifying information (PII) or offensive content.

Our dataset is built by repurposing existing VLN resources grounded in Matterport3D indoor environments and by automatically generating navigation dialog from those trajectories. The resulting data consists of navigation paths, scene-grounded object references, and generated question--answer pairs about indoor environments. It is not designed to include names, personal identities, contact information, or other uniquely identifying information about individuals.

\section{Use of AI Assistants}
\label{sec:ai-assistants}

AI assistants were used in two ways in this work. First, they were used in the research pipeline for automatic data generation, including dialog reformatting and caption-based synthesis, as described in the main paper. Second, they were used during manuscript preparation for writing support, such as grammar correction and improving clarity and flow. All scientific decisions, experimental design, interpretations, and final wording were reviewed and verified by the authors.

\begin{figure*}[t]
\begin{subfigure}[b]{\linewidth}
    \begin{tcolorbox}[colframe=gray!50!black, colback=gray!10!white,
                  title=Prompt for Simple Panoramic Caption,
                  width=\textwidth, sharp corners=southwest]
    {\ttfamily \scriptsize \linespread{1.2}\selectfont
    Generate a sentence that describes the given panoramic image. 
    Start your question with "I'm in ..." or "I can see ...".
    }
\end{tcolorbox}
    \vspace{-0.2cm}
\caption{Prompt for Simple Panoramic Caption}
\end{subfigure}

\begin{subfigure}[b]{\linewidth}
\begin{tcolorbox}[colframe=gray!50!black, colback=gray!10!white,
                  title=Prompt for Region Grounded Caption,
                  width=\textwidth, sharp corners=southwest]
{\ttfamily \scriptsize \linespread{1.2}\selectfont
Please describe the \{region\} in this image.
}
\end{tcolorbox}
    \vspace{-0.2cm}
\caption{Prompt 
for region grounded caption}
\end{subfigure}

\begin{subfigure}[b]{\linewidth}
\begin{tcolorbox}[colframe=gray!50!black, colback=gray!10!white,
                  title=Prompt for Object Caption,
                  width=\textwidth, sharp corners=southwest]
{\ttfamily \scriptsize \linespread{1.2}\selectfont
Please describe the \{obj\} in this image. If there's no \{obj\}, response 'None'
}
\end{tcolorbox}
    \vspace{-0.2cm}
\caption{Prompt for object caption}
\end{subfigure}

\begin{subfigure}[b]{\linewidth}
\begin{tcolorbox}[colframe=gray!50!black, colback=gray!10!white,
                  title=Prompt for Region and Object Grounded Caption,
                  width=\textwidth, sharp corners=southwest]
{\ttfamily \scriptsize \linespread{1.2}\selectfont
You will be provided with a panoramic image caption describing an entire space and detailed captions of key objects visible in that image. Based on the input, describe the surrounding environment clearly and concisely. Your description should include the structure, layout, and the locations of key objects. Do not add anything, but just the sentence describing around.
}
\end{tcolorbox}
    \vspace{-0.2cm}
\caption{Prompt for region and object grounded caption}
\end{subfigure}

\caption{Prompts used to generation captions for navigation questions.}

\label{fig:prompt_q_panoramic}
\end{figure*}

\begin{figure*}[t]
\centering

\begin{subfigure}[t]{\textwidth}
\centering
\begin{tcolorbox}[colframe=gray!50!black, colback=gray!10!white,
                  title=Prompt for Goal-conditioned Reformat,
                  width=\textwidth, sharp corners=southwest]
{\ttfamily \scriptsize \linespread{1.2}\selectfont
You will be provided with a sequence of sentences describing a path.

Your task is to rewrite the path description in two versions:

1. with\_goal: A version that includes reaching the final destination

2. without\_goal: A version that excludes any mention of reaching the destination

Both versions should:

- Be concise while preserving key spatial references

- Focus on objects, rooms, and layout

- Use directions only when necessary

- Remove terms implying current situation (Start facing ~, You're in a ~, You're facing)

For with\_goal version, Always mention that you've arrived at the destination at the end of the description with expressions like:
That's the goal, You have arrived, That's your destination, The goal, You've reached your destination, This is the final spot, You're at the destination, Mission accomplished, End point reached, You've made it.

For without\_goal version, Remove any mentions of arriving at or reaching the final destination.
}
\end{tcolorbox}
\caption{Prompt template for goal-conditioned path reformatting.}
\end{subfigure}

\vspace{0.4cm}

\begin{subfigure}[t]{\textwidth}
\centering
\begin{tcolorbox}[colframe=gray!50!black, colback=gray!10!white,
                  title=Prompt for Natural Dialog Reformat,
                  width=\textwidth, sharp corners=southwest]
{\ttfamily \scriptsize \linespread{1.2}\selectfont
You will rewrite a dialog between a Navigator and a Guide in an indoor navigation task.

TASK CONTEXT:

- The Navigator and Guide work together to reach a goal room.

- Both Navigator and Guide know about a shared object in the goal room.

- The Guide knows the goal room but the Navigator does not.

- The object may appear in other rooms, so the Navigator must ask clarifying questions.

- The Guide cannot see the Navigator's position but knows the environment layout.

- Dialog alternates turns: Navigator → Guide → Navigator … (no limit on turns).

- Dialog always starts with the Navigator.

OBJECTIVE:

- Rephrase the dialog to sound like natural conversation.

- Style:

  * Use casual, friendly language ("Ok", "Alright", "Got it").

  * Add acknowledgments ("I see", "Understood").

  * Smooth transitions between turns.

  * Vary vocabulary, keep meaning the same.

- Actively incorporate mentions of the shared object in the conversation.

- The Guide's final response should mention reaching the goal room.

RULES:

1. Keep the same number of turns and same speakers (Navigator/Guide).

2. Do not add or remove turns.

3. Do not add new objects, rooms, or details not in input.

4. Never use phrases like "You're looking at…" or "You are facing…".
   - Instead: ask questions ("Do you see…?") or omit.

5. Do not drop or shorten navigation instructions.

6. Preserve all key details:

   - Objects and attributes (color, shape, material).

   - Room names and types.

   - Spatial relations and directions.

7. Do not use em dashes (—). Use commas, periods, or "and".

OUTPUT FORMAT:

Return ONLY valid JSON:

{

  "reformatted": [

    { "Navigator": "...", "Guide": "..." },

    { "Navigator": "...", "Guide": "..." }

  ]

}

No extra text, no markdown, no explanations.
}
\end{tcolorbox}
\caption{Prompt template for natural dialog reformatting.}
\end{subfigure}
\caption{Prompt for dialog reformatting.}
\label{fig:prompt_dialog_reformat}
\end{figure*}

\begin{figure*}[t]
\centering

\begin{subfigure}{\textwidth}
 \centering
 \scriptsize
 \renewcommand{\arraystretch}{1.25}

 \begin{tabularx}{\linewidth}{@{}X@{}}
 \toprule
 \textbf{Example} \\
 \midrule

 \textbf{Original Instruction} \\
 ``We start off looking at the corner of a wall. As you turn around you see a large open are with doorways to many rooms around you. On one of the walls you see a white rectangular box. Take a step towards that white rectangular shape and turn to look down the hall and you will see two chairs directly in front of you down the hall. And before those two chairs on the right there is an open walkway. Take a step towards that open walkway on your right. You will be looking into a room that has brown cabinets and a shelf with a white pillow and radio looking thing on it and you are done.'' \\[4pt]

 \textbf{Reformatted (without goal)} \\
 ``From the corner of a wall, turn to see a large open area with doorways to various rooms. Approach the white rectangular box on the wall, then look down the hall to see two chairs ahead. To the right of the chairs, there’s an open walkway. Step into the walkway to view a room with brown cabinets and a shelf containing a white pillow and a radio.'' \\[4pt]

 \textbf{Reformatted (with goal)} \\ 
 ``Turn away from the wall to find a large open area with doorways to various rooms. Move toward the white rectangular box on the wall, then look down the hall to see two chairs ahead. To the right of those chairs is an open walkway; step into it. You’ll see a room with brown cabinets and a shelf containing a white pillow and a radio. You've reached your destination.'' \\
 \bottomrule
 \end{tabularx}

 \caption{Instruction Refining.}
 \label{fig:goal_branching_examples_sub}
\end{subfigure}

\vspace{1.2em}

\begin{subfigure}{\textwidth}
 \centering
 \scriptsize

 \resizebox{\textwidth}{!}{%
 \begin{tabular}{@{}>{\bfseries}m{0.03\textwidth} p{0.47\textwidth} p{0.47\textwidth}@{}}
 \toprule
 & \textbf{Before Reformat} & \textbf{After Reformat} \\
 \specialrule{1pt}{2pt}{2pt}

 Q1 &
 I'm in a long hallway filled with books from floor to ceiling. The bookshelf stretches the entire length of the hallway, packed with books of all sizes and types, creating an impressive display of knowledge and resources. &
 I'm in this long hallway packed with books all the way up to the ceiling. The bookshelf runs the entire length of the hallway, filled with all sorts of books, creating an amazing sight. \\
 \addlinespace[4pt]

 A1 &
 You are standing in a library, to the right you have a book shelf, turn back from the place where you are standing, walk straight, till the end, turn right, walk straight, until you reach the book shelf in-front of you, turn right, walk straight, walk straight until you reach the book shelf in-front of you and a staircase on your right, get down the stairs completely. Slightly turn right, in-front of you there is a large table with chairs around it, to the right you have a book shelf, stand between the table and book shelf, turn back. Sorry, turn right, in-front of you there is an open door walk to the open door, in-front of you there is a staircase, stand on the staircase, that's your end point. &
 Awesome! In the library, head straight to the end and take a right. Keep going until you reach another bookshelf, then make another right and head toward the bookshelf with a staircase on its right. Go down the staircase completely, and then turn a bit right to find a big table surrounded by chairs and another bookshelf on the right—stand right between those two. Next, turn right toward the open door and walk through it to reach the staircase. \\

 Q2 &
 The library in this image is a room with a stone archway and a white wall. The room appears to be a part of a university, as it is described as a ``college library.'' The stone archway adds a unique architectural element to the space, making it visually appealing. The white wall complements the stone archway, creating a clean and elegant atmosphere. The room is likely used for studying, research, and quiet contemplation, as it is a library setting. &
 I see the library has a stone archway and a white wall. It feels like a university space, really charming. \\
 \addlinespace[4pt]

 A2 &
 You are now facing the white color door, from there take a left turn. Start going down the stairs and stand on the tenth step, you will find your destination point. &
 Perfect! Now, turn left from the white door and go down the stairs until you hit the tenth step. That’s your goal. \\
 \bottomrule
 \end{tabular}
 }

 \vspace{1.2em}

 \resizebox{\textwidth}{!}{%
 \begin{tabular}{@{}>{\bfseries}m{0.03\textwidth} p{0.47\textwidth} p{0.47\textwidth}@{}}
 \toprule
 & \textbf{Before Reformat} & \textbf{After Reformat} \\
 \specialrule{1pt}{2pt}{2pt}

 Q1 &
 I am in a large, wooden wardrobe with a mirrored door. The closet is filled with various clothing items, including shirts and suits, and is well-organized, making it easy to find and access the clothes stored inside. &
 I'm inside this large wooden wardrobe with a mirrored door. It's packed with all kinds of clothes like shirts and suits, and everything is organized really well, so it's easy to find what I need. \\
 \addlinespace[4pt]

 A1 &
 Walk down the hallway to your right and move past the bed. Walk through the door on the right of the painting of thr girl and wait there. &
 Alright! Start by heading down the hallway to your right and walk past the bed. Then, go through the door on the right side of the girl’s painting and just wait there. \\

 Q2 &
 I'm in a curved hallway with a staircase and glass railing, surrounded by various artworks on the walls, including paintings and photographs. &
 I've made it to a curved hallway that has a staircase and a glass railing. The walls are decorated with various artworks, including paintings and photos. \\
 \addlinespace[4pt]

 A2 &
 down more stairs, and into the little room left of the silver mirrored planter &
 Nice! Now, head down those stairs and enter the small room to the left of the silver mirrored planter. You've finally arrived at your destination. \\
 \bottomrule
 \end{tabular}
 }

 \caption{Conversational Smoothing (Q: Question, A: Answer).}
 \label{fig:dialog_reformat_2samples_sub}
\end{subfigure}

\caption{\textbf{Example of dialog reformatting.}
(a) Examples of \textbf{Step 1 (Instruction Refining)}, where raw VLN instructions are reformatted into `without goal` and `with goal`.
(b) Examples of \textbf{Step 2 (Conversational Smoothing)}, where the sequence of captions and refined instructions is paraphrased into fluent dialog.}
\label{fig:dialog_reformat_combined}

\end{figure*}

\begin{table*}[ht]
\centering
\footnotesize

\begin{subtable}{\linewidth}
\centering
\label{tab:qualitative_questions}
\begin{tabularx}{\linewidth}{@{} X X X@{}}
\toprule
\textbf{Ground Truth (GT)} & \textbf{RAIN (Baseline)} & \textbf{RAINbow (Ours)} \\
\midrule
Hi I'm on the \textbf{top floor}. to my left I see a \textbf{wooden desk}, to my right a \textbf{wooden cabinet} under a \textbf{tapestry}. If I walk past the tapestry there is a \textbf{living room area} with a \textbf{brown couch, armchair, TV and a chest}. & 
i 'm in a \textcolor{blue}{living room} with a \textcolor{blue}{fireplace} , \textcolor{red}{a fireplace , and a fireplace .} &
i 'm in this cozy \textcolor{blue}{living room} with a wooden floor, and there 's a \textcolor{blue}{wooden table} right in the middle . it 's got a warm vibe with a lamp on the table next to the bed . i can see a white chair with a purple seat in front of the window .\\
\addlinespace[1pt] \midrule \addlinespace[1pt]

Hi, I am at a \textbf{gym place} full of \textbf{different gym equipments}. If there isn't any other gym room in the house, it should be easy to find out my current location. Otherwise, I will find more distinguishable features of the room. Thank you for helping me find the goal room. & 
i 'm in a \textcolor{blue}{gym room} with a lot of . \textcolor{red}{there are two black chairs and a water dispenser .} & 
i 'm in a big , open \textcolor{blue}{gym} with a bunch of \textcolor{blue}{exercise machines}. there 's a treadmill on the left and a bench in the middle . i also see a large , sliding glass door to the outside .\\
\bottomrule
\end{tabularx}
\caption{Generated Question}
\end{subtable}

\vspace{1em}

\begin{subtable}{\linewidth}
\centering
\label{tab:qualitative_answers}
\begin{tabularx}{\linewidth}{@{} X X X@{}}
\toprule
\textbf{Ground Truth (GT)} & \textbf{RAIN (Baseline)} & \textbf{RAINbow (Ours)} \\
\midrule
Come out of the \textbf{bathroom} and you'll see a big \textbf{bedroom} with a \textbf{big bed} and lots of \textbf{sofas}. The bed has white sheets and \textbf{blue pillows}. Come out of the bedroom through the door that's left to the \textbf{Tv}. Next to the door is a\textbf{ picture of a sailboat}. Once you've reached the hallway that has some writings hanging on the wall to your left, let me know.
& ok , i found you . please walk out of the \textcolor{blue}{bathroom} and go straight until you see a \textcolor{blue}{bedroom} on your left . \textcolor{red}{turn right and go straight until you see a bedroom on your left . turn right and go straight until you see a bedroom on your left .} go up the stairs and let me know 
& alright , head out of the \textcolor{blue}{bathroom} through the open arch and walk straight to the \textcolor{blue}{bedroom} entrance . go through the doors , turn left to find \textcolor{blue}{the lounge area in the center} . look slightly right to spot a painting of a ship on the left wall . then , go through the open door in front of you and walk toward the \textcolor{blue}{photo frame on the front wall .} turn right and head straight to the staircase on your left , climbing up . \\
\addlinespace[1pt] \midrule
\addlinespace[1pt]

Okay, go to the \textbf{dining table} and walk through the \textbf{right side} of the dining table. You'll see a room both in left and right. \textbf{Enter the right one}. Then, you'll see a \textbf{stair on your right side}. Can you see this stair? & 
ok , you 're in the \textcolor{blue}{dining room} . \textcolor{red}{you have to go to the dining room .} & 
no problem ! from the \textcolor{blue}{dining table} , turn around and head towards the wall . slightly turn left and go to the open door . exit the room and turn right to find a \textcolor{blue}{staircase}. go up the steps , then turn right again to see another open door . enter that room and stand in front of the mirror . you 've made it to the bathroom !\\
\bottomrule
\end{tabularx}
\caption{Generated Answer}
\end{subtable}

\caption{\textbf{Qualitative Comparison of Generated Dialog.} Comparison of Ground Truth (GT) questions and answers versus models trained on RAIN and RAINbow data. Bold text indicates essential content in the Ground Truth (GT), \textcolor{blue}{blue text} denotes overlapping content found in the generated outputs, and \textcolor{red}{red text} highlights non-fluent content. RAINbow significantly improves fluency and reduces redundancy compared to the RAIN baseline.}
\label{tab:qualitative_questions_and_answers}
\end{table*}

\end{document}